\PassOptionsToPackage{square,comma,numbers,sort&compress,super}{natbib}
\documentclass{article}

\usepackage[preprint]{neurips_2019}

\usepackage[utf8]{inputenc} % allow utf-8 input
\usepackage[T1]{fontenc}    % use 8-bit T1 fonts
\usepackage{hyperref}       % hyperlinks
\usepackage{url}            % simple URL typesetting
\usepackage{booktabs}       % professional-quality tables
\usepackage{amsfonts}       % blackboard math symbols
\usepackage{nicefrac}       % compact symbols for 1/2, etc.
\usepackage{microtype}      % microtypography

\usepackage{amsmath,amsthm,amssymb,commath,bm,bbm}
\newtheorem{theorem}{Theorem}
\newtheorem{proposition}{Proposition}
\newtheorem{definition}{Definition}
\usepackage{graphicx}
\usepackage{subfig}
\usepackage{shortbold}

\title{On the Learning Dynamics of Two-layer Nonlinear Convolutional Neural Networks}

\author{
  Bing Yu\thanks{Equal contribution}\\
  Peking University\\
  \texttt{byu@pku.edu.cn}\\
  \And
  Junzhao Zhang\footnotemark[1]\\
  Peking University\\
  \texttt{1801213960@pku.edu.cn}\\
  \And
  Zhanxing Zhu\\
  School of Mathematical Science, Peking University\\
  Center for Data Science, Peking University\\
  Beijing Institute of Big Data Research\\
  \texttt{zhanxing.zhu@pku.edu.cn}\\
}

\begin{document}

\maketitle

\begin{abstract}
  Convolutional neural networks (CNNs) have achieved remarkable performance in various fields, particularly in the domain of computer vision. However, why this architecture works well remains to be a mystery. In this work we move a small step toward understanding the success of CNNs by investigating the learning dynamics of a two-layer nonlinear convolutional neural network over some specific data distributions. Rather than the typical Gaussian assumption for input data distribution, we consider a more realistic setting that each data point (e.g. image) contains a specific pattern determining its class label. Within this setting, we both theoretically and empirically show that some convolutional filters will learn the key patterns in data and the norm of these filters will dominate during the training process with stochastic gradient descent. And with any high probability, when the number of iterations is sufficiently large, the CNN model could obtain $100\%$ accuracy over the considered data distributions. Our experiments demonstrate that for practical image classification tasks our findings still hold to some extent. 
\end{abstract}

\section{Introduction}

Convolutional neural networks (CNNs) have recently exhibited great performance in various fields such as computer vision\cite{krizhevsky2012imagenet}, natural language processing~\cite{dauphin2017language} and reinforcement learning~\cite{silver2016mastering}. But why CNNs work so well still remains mysterious. 

Recently, some research works attempted to understand CNNs from different perspectives. From representational view, \cite{lecun2015deep,anselmi2016unsupervised} pointed out that conceptually deep convolution networks are computing
progressively more powerful invariants as depth increases. \cite{mallat2016understanding} further showed that the network filters are guiding nonlinear contractions to reduce the data variability in directions of local symmetries. Some works tried to visualize the  covonlutional filters or feature maps learned by CNNs to see what patterns the CNNs have captured~\cite{zeiler2014visualizing,mordvintsev,olah2017feature}. Due to the high complexity of deep CNNs, most of theoretical analysis lies in investigating the two-layer settings. When considering the learning dynamics or the generalization bias of CNNs,  existing works either assumed Gaussian distribution for input data or only studied linear convolutional networks~\cite{goel2018learning,zhong2017learning,brutzkus2017globally,du2018gradientCNN,wu2018towards,gunasekar2018implicit}.

In this work, we move a further step along the direction of theoretically analyzing CNNs.  We investigate the learning dynamics of a two-layer nonlinear CNN using stochastic gradient descent (SGD) for binary classification, with a more realistic assumption on the data distribution. In particular, an input data point is labeled positive if and only if there is a specific pattern in it and is labeled negative if and only if there exists another different pattern. This assumption is motivated by image classification tasks, where an image belongs to a object class if and only if this image owns some specific patterns (e.g. an image of vehicle has a wheel pattern). We emphasize that analysis under this setting could potentially help us to understand the learning mechanism of CNNs.

We name those patterns that play an important role in classification as key patterns, and others as non-key patterns. We prove that, under some assumptions over data distribution, some filters of the two-layer nonlinear CNN will learn the direction of the key patterns and the norm of them will increase much faster than others. Therefore, these filters that learn the key patterns will dominate during the training, and then with any large probability (the randomness is over the initialization and the selection of training data in each iteration), after enough number of iterations, good generalization accuracy can be guaranteed. To the best of our knowledge, we are the first to analyze the learning dynamics of two-layer nonlinear CNNs in  more realistic settings. Our theoretical findings that the covonlutional filters could capture the key pattern in data during training shed light on understanding the underlying mechanisms of CNNs.

We also conduct experiments over some practical datasets that does not fully satisfy our assumptions. And we observe our theoretical findings still hold that there exist some filters learn the key patterns in data and dominate.

\subsection{Related Works}
We provide  some previous works on the learning dynamics of fully connected networks (FCNs) and CNNs.

\textbf{Trajectory-based Analysis of FCNs.}  Recently, there are many works analyze the convergence property of FCNs through a trajectory-based analysis~\cite{li2018learning,allen2018convergence,zou2018stochastic,du2018gradient_a,du2018gradient_b,brutzkus2017sgd}. These works mainly focused on over-parameterized two-layer FCNs. According to~\cite{arora2019fine}, their proof techniques can be roughly classified into two categories. \cite{li2018learning,allen2018convergence,zou2018stochastic} analyzed the trajectory of parameters and showed that on the trajectory, the objective function satisfies certain gradient dominance property, while \cite{du2018gradient_a,du2018gradient_b} analyzed the trajectory of network predictions on training samples and showed that it enjoys a strongly-convex-like property.

\textbf{Learning dynamics of CNNs.}   The works~\cite{goel2018learning,zhong2017learning,brutzkus2017globally,du2018gradientCNN} focused on analyzing the learning a two-layer CNN for regression problem with Gaussian inputs, under the teacher-student network settings. They showed that the dynamics could converge to true parameter or zero loss. \cite{wu2018towards} considered learning a two-layer CNN for some binary classification task assuming a restricted input distribution. Multi-layer linear CNNs were studied in \cite{gunasekar2018implicit} for binary classification problem with linear separable data distribution. And they proved that the dynamics could converge to a linear predictor related to the $l_{2/L}$ bridge penalty in the frequency domain.

%\subsection{Contribution}
%\begin{itemize}
    %\item To the best of our knowledge, we are the first to analyze the learning dynamcs of two-layer nonlinear CNNs in  more realistic settings. Our theoretical findings that the covonlutional filters could capture the key pattern in data during training shed light on understanding the underlying mechanisms of CNNs.
    %\item 
%\end{itemize}

\section{Problem Setup}\label{setup}
\subsection{Data Distribution and Network Architecture}
\paragraph{Binary classification task} We consider using CNNs to solve a binary classification problem, and the input data $x \in \Rbb^D$, the convolutional filters of the CNN $w \in \Rbb^d$ and there exists a positive integer $k$ such that $D=kd$. We consider the following two data distributions.

\begin{definition}\label{defclean}
\textbf{(The Clean Distribution $\Dcal$)} Let $p_+,p_-\in\{p\in \Rbb ^d:{\|p\|}_2=1\}$ be the positive key pattern and negative key pattern~($p_+\ne p_-$), and let $(x,y)$ be a sample drawn from $\mathcal{D}$, where $x\in  \Rbb^D$, $y\in\{-1,1\}$, then

(1) $P\{y=1\}=P\{y=-1\}=\frac{1}{2}$;

(2) If $y=1$,  there exists $u^*\in[k]$, s.t. $x_{u^*d+1,\dots,(u^*+1)d}=p_+$, and for any $u\in[k]$, $u\ne u^*$, $x_{ud+1,\dots,(u+1)d}$ is an i.i.d. sample drawn from $U(\{p\in \Rbb^d:\langle p_+,p\rangle=0,\langle p_-,p\rangle=0,{\|p\|}_2=1\})$, where $U(S)$ is the uniform distribution over set $S$;

(3) If $y=-1$, there exists $u^*\in[k]$, s.t. $x_{u^*d+1,\dots,(u^*+1)d}=p_-$, and for any $u\in[k]$, $u\ne u^*$, $x_{ud+1,\dots,(u+1)d}$ is an i.i.d. sample drawn from $U(\{p\in \Rbb^d:\langle p_+,p\rangle=0,\langle p_-,p\rangle=0,{\|p\|}_2=1\})$.
\end{definition}

We then define another data distribution by adding some noise to the non-key patterns of data point drawn from $\Dcal$, as Definition~\ref{defnoisy} shows.

\begin{definition}\label{defnoisy}
\textbf{(The $\epsilon$-Noisy Distribution ${\Dcal}_{\epsilon}$)} Let $p_+,p_-\in\{p\in \Rbb^d:{\|p\|}_2=1\}$ be the positive key pattern and negative key pattern~($p_+\ne p_-$), and let $(x,y)$ be a sample drawn from $\Dcal_{\epsilon}$, where $x\in \Rbb ^D$, $y\in \{-1,1\}$, then $x$ can be decomposed by
\begin{equation}
x=x_0+x_1+x_2,
\end{equation}
where

(1) $(x_0,y)$ is a sample drawn from the clean distribution $\Dcal$;

(2) $x_{1,u^*d+1...(u^*+1)d}=x_{2,u^*d+1...(u^*+1)d}=0$;

(3) For any $u\in[k],u\ne u^*$, $x_{1,ud+1...(u+1)d}$ is a sample drawn from $U(\{p\in \Rbb ^d:p\in span\{p_+,p_-\},{\|p\|}_2\le\epsilon\})$, $x_{2,ud+1...(u+1)d}$ is a sample draw from $\{p\in \Rbb ^d:\langle p_+,p\rangle=0,\langle p_-,p\rangle=0,{\|p\|}_2\le\epsilon\}$.
\end{definition}

It is obvious that the clean distribution $\Dcal$ is linearly separable, but when $k$ is sufficiently large, the $\epsilon$-noisy distribution ${\Dcal}_{\epsilon}$ is not linearly separable any more, as Proposition~\ref{nonls} shows.

\begin{proposition}\label{nonls}
Assumed that $p_+=-p_-=p$, then when $k\ge\frac{2}{\epsilon}+1$, the $\epsilon$-noisy distribution ${\Dcal}_{\epsilon}$ is not linearly separable.
\end{proposition}
The proof is left to Appendix. For simplicity we only prove the $p_+=-p_-$ case. But for general cases similar results can be obtained.

\paragraph{CNN architecture} In this work, we consider the following two-layer nonlinear CNN architecture
\begin{equation}
F(x;W,a)=\sum_{i=1}^h a_if(x;w_i),
\end{equation}
where
\begin{equation}
f(x;w)=\max\{0,\langle x_{1...d},w\rangle,\langle x_{d+1...2d},w\rangle,...,\langle x_{D-d+1...D},w\rangle\}
\end{equation}

where $x={(x_1,...,x_D)}^T\in \Rbb ^D$ is input data, and $x_{ud+1...(u+1)d}={(x_{ud+1},x_{ud+2},...,x_{(u+1)d})}^T$ is a pattern of $x$. $w_1,...,w_h\in \Rbb^d$ are filters of the convolutional layer, the scalars $a_1,...,a_h$ are weights of the second layer, $W=\{w_1,...,w_h\}$ stands for the set of parameters in the first layer, $a=\{a_1,...,a_h\}$ stands for the set of parameters in the second layer. Note that $f$ is a compositional function of convolution, ReLU activation and max pooling.   

This two-layer CNN architecture can be interpreted as: for input data $x$, conducting a  length-$d$, stride-$d$ convolution followed by ReLU activation and size-$k$, stride-$k$ max pooling in the first layer, then a linear combination in the second layer.

\subsection{Learning Process} 
To simplify the analysis, we only learn the CNN's filters $w_i,i=1,\dots,h$. And we assume that $a_i$ equals to $1$ or $-1$ with equal probability $\frac{1}{2}$ and is fixed during training. We then use $F(x;W)$ to replace $F(x;W,a)$. We employ 
a variant of hinge loss as our learning target, 
\begin{equation}
\ell(F(x;W),y)=-yF(x;W).
\end{equation}
The learning process tries to minimize the following loss function under clean or noisy distribution,
\begin{equation}
{\min}_{W}E_{(x,y)\sim \Dcal}[\ell(F(x;W),y)] \text{ or }  {\min}_{W}E_{(x,y)\sim \Dcal_{\epsilon}}[\ell(F(x;W),y)].
\end{equation}
Note that the two objectives are non-convex and non-concave, we can show it by providing a concrete example, presented in  Proposition~2 of Appendix.

% The proof is leaving to appendix.
SGD with batch size one is employed for solving the optimization problem. At $(t+1)$-th iteration, for each filter $w_i$, $i=1,\dots,h$, the update is 

\begin{align}
w_i^{(t+1)}&=w_i^{(t)}+\eta  a_iy^{(t+1)}x_{ud+1...(u+1)d}^{(t+1)}\Ibb_{\{f(x^{(t+1)};w_i^{(t)})>0\}}\\
u&={\argmax}_v\{\langle x_{vd+1...(v+1)d}^{(t+1)},w_i^{(t)}\rangle\}\label{dynamics}
\end{align}

where $\eta$ is the learning rate, $(x^{(t+1)},y^{(t+1)})$ is i.i.d drawn from the training data distribution (which will be specified in Section~\ref{mainresults}), the initialization of $w_i$ is $w_i^{(0)}\sim U(\{w\in \Rbb^d:{||w||}_2=1\})$. And we say ``$x_{ud+1...(u+1)d}^{(t+1)}$ is activated by $w_i^{(t)}$'' or ``$x_{ud+1...(u+1)d}^{(t+1)}$ is activated'' if $u={\argmax}_v\{\langle x_{vd+1...(v+1)d}^{(t+1)},w_i^{(t)}\rangle\}$.

\section{Main Results}\label{mainresults}
We first consider learning the population loss. Under different data distributions, either clean or noisy one, we show that the convolutional filters can learn the direction of key patterns in data, described in Theorem~\ref{maxclean} and \ref{maxnoisy}, respectively.  Secondly, the case of training with empirical data is studied. Its learning dynamics and generalization performance are characterized in Theorem~\ref{limiteddata}.   
The proofs are elaborated in Appendix.

\subsection{Learning with Population Loss}

\begin{figure}[ht]
\vspace{-7mm}
    \centering
    \subfloat[\label{obuse}]{\includegraphics[width=0.3\textwidth]{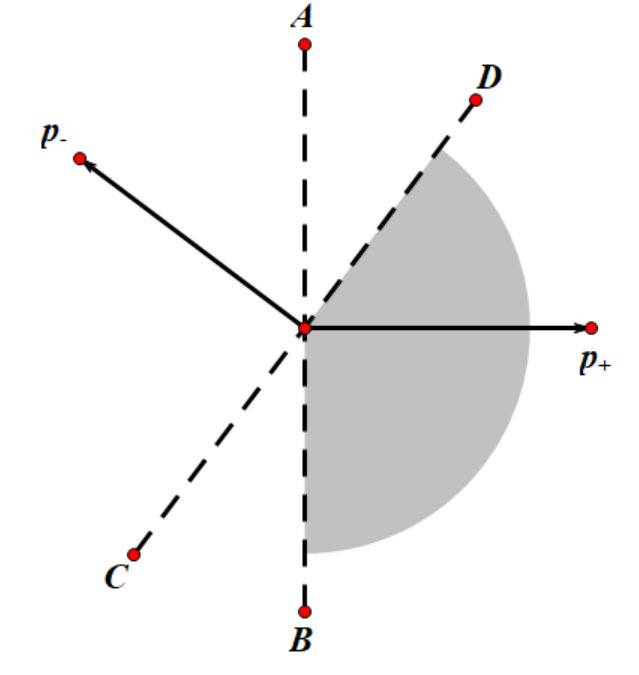}}\hfill
    \subfloat[\label{sharp}]{\includegraphics[width=0.3\textwidth]{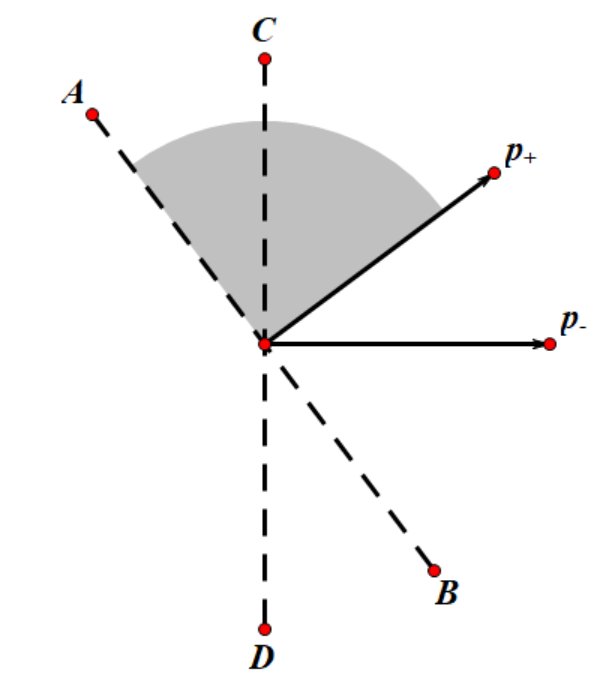}}\hfill
    \vspace{-2mm}
    \caption{Illustration for Theorem~\ref{maxclean}. (\textbf{a}) Theorem~\ref{maxclean}(2), line segments $AB$ and $CD$ satisfy that $AB\perp p_+$, $CD\perp p_-$. (\textbf{b}) Theorem~\ref{maxclean}(3),  $AB$ and $CD$ satisfy that $AB\perp p_+$, $CD\perp p_-$.}
    \label{obusesharp}
\end{figure}

\begin{theorem}\label{maxclean}
(\textbf{Learning with Clean Data Distribution}) 
Assumed that $(x^{(t+1)},y^{(t+1)}),t=0,1,...$ are i.i.d. randomly drawn from the clean distribution $\Dcal$. Then

(1) When $p_+=-p_-=p$,

(1.1) For any $\sigma\in(0,\frac{1}{2})$, let ${\theta}_t$ be the angle between $a_ip$ and $w_i^{(t)}$, then
\begin{equation}
   P\{\sin{\theta}_t\le \mathcal{O}(t^{-(\frac{1}{2}-\sigma)})\}\ge1-\mathcal{O}(t^{-\sigma}) 
\end{equation}

(1.2) For any $\sigma\in(0,\frac{1}{2})$, $\epsilon>0$, when
\begin{equation}
t\ge max\{\frac{(2^{k+2}-4)(h+2)}{\epsilon},{(\frac{(\eta+2)(h+2)}{\epsilon})}^{\frac{1}{\sigma}},{(\frac{2^{k+1}(2h+1)}{\eta})}^{\frac{2}{1-2\sigma}}\},
\end{equation}
with probability at least $1-{(\frac{1}{2})}^{h-1}-\epsilon$, after $t$ steps, the classification accuracy over $\Dcal$ will be $1$;

(2) When $\langle p_+,p_-\rangle\le0$ and $p_+\ne-p_-$,

(2.1) If $a_i=1$, let ${\theta}_t$ be the angle between $p_+$ and $w_i^{(t)}$, otherwise let ${\theta}_t$ be the angle between $p_-$ and $w_i^{(t)}$.
Decomposed $w_i^{(t)}$ by $w_i^{(t)}=w_{\parallel,i}^{(t)}+w_{\perp,i}^{(t)}$, where $w_{\parallel,i}^{(t)}\in span\{p_+,p_-\}$, $w_{\perp,i}^{(t)}$ is perpenticular to $p_+$ and $p_-$.
If $w_{\parallel,i}^{(0)}$ is in the shaded region of Figure~\ref{obuse}, then for any $\sigma\in(0,\frac{1}{2})$,
\begin{equation}
   P\{\sin{\theta}_t\le \mathcal{O}(t^{-(\frac{1}{2}-\sigma)})\}\ge1-\mathcal{O}(t^{-\sigma}) 
\end{equation}

(2.2) For any $\sigma\in(0,\frac{1}{2}),\epsilon>0$, when
\begin{equation}
t\ge max\{\frac{(2^{k+2}-4)(h+2)}{\epsilon},{(\frac{(\eta+2)(h+2)}{\epsilon})}^{\frac{1}{\sigma}},{(\frac{2^{k+1}(2h+1)}{\eta})}^{\frac{2}{1-2\sigma}}\},
\end{equation}
with probability at least $1-2\cdot{(\frac{7}{8})}^h-\epsilon$, after $t$ steps the classification accuracy over $\Dcal$ will be $1$;

(3) When $\langle p_+,p_-\rangle>0$, $p_+\ne p_-$, positive and negative examples are sampled alternatively,

(3.1) Let $T$ be a positive integer. If $a_i=1$, let ${\theta}_t$ be the angle between $p_+-\langle p_+,p_-\rangle p_-$ and $w_i^{(T+t)}$, otherwise let ${\theta}_t$ be the angle between $p_--\langle p_+,p_-\rangle p_+$ and $w_i^{(T+t)}$.
Decomposed $w_i^{(t)}$ by $w_i^{(t)}=w_{\parallel,i}^{(t)}+w_{\perp,i}^{(t)}$, where $w_{\parallel,i}^{(t)}\in span\{p_+,p_-\}$, $w_{\perp,i}^{(t)}$ is perpenticular to $p_+$ and $p_-$.
If $T$ is sufficiently large and $w_{\parallel,i}^{(T)}$ is in the shaded region of Figure~\ref{sharp}, then for any $\sigma\in(0,\frac{1}{2})$,
\begin{equation}
P\{\sin{\theta}_{t+T}\le \mathcal{O}(t^{-(\frac{1}{2}-\sigma)})\}\ge1-\mathcal{O}(t^{-\sigma})-\mathcal{O}(T^{-\sigma})
\end{equation}

(3.2) For any $\sigma\in(0,\frac{1}{2}),\epsilon>0$, when $T$ and $t$ are sufficiently large, then with probability at least $1-2\cdot{(\frac{7}{8})}^h-\epsilon$, after $T+t$ steps the classification accuracy over $\Dcal$ will be $1$.
\end{theorem}

Theorem~\ref{maxclean} implies that when training data are i.i.d. drawn from $\Dcal$, some filters of the considered CNN will approximately learn the direction of the key pattern. According to our proof, the norm of these filters increase much faster than that of other filters during training process, so these filters will be dominant when testing. And with the increasing number of filters, the probability of achieving better classification accuracy can be guaranteed. 

Similar results can also be obtained over the noisy distribution ${\Dcal}_{\epsilon}$. For simplicity, we only present the $p_+=-p_-$ case, as Theorem~\ref{maxnoisy} shows.

\begin{theorem}\label{maxnoisy}
(\textbf{Population Loss with Noisy Data Distribution})
Assumed that $p_+=-p_-=p$,  $(x^{(t+1)},y^{(t+1)}),t=0,1,...$ are i.i.d. drawn from the $\epsilon$-noisy distribution $\mathcal{D}_{\epsilon}$, where $\epsilon$ satisfies $0<\epsilon<2^{-2k-1}$. Then

(1) For any $\sigma\in(0,\frac{1}{2})$, let ${\theta}_t$ be the angle between $a_ip$ and $w_i^{(t)}$, then
\begin{equation}
  P\{\sin{\theta}_t\le \mathcal{O}(t^{-(\frac{1}{2}-\sigma)})\}\ge1-\mathcal{O}(t^{-\sigma})
\end{equation}

(2) If $|\{i|a_i=1\}|=|\{i|a_i=-1\}|$, the classification accuracy over $\Dcal_{\epsilon}$ will be 1 when $t\to\infty$.
\end{theorem}

According to Proposition~\ref{nonls}, $\Dcal_{\epsilon}$ is not linearly separable when $k$ is sufficiently large. So Theorem~\ref{maxnoisy} shows that with any large probability, after enough training steps over $\Dcal_{\epsilon}$, our considered CNN architecture can obtain accuracy 1 over this non-linearly separable data distribution.

\subsection{Learning with Empirical Loss}
We can also prove similar results when training data is limited, drawn from population distribution. But for simplicity, we only prove the $p_+=-p_-$ case, as Theorem~\ref{limiteddata} describes.

\begin{theorem}\label{limiteddata}
(\textbf{Empirical Loss with Clean Data Distribution})
Assumed that $p_+=-p_-=p$, $(x^{(t+1)},y^{(t+1)}),t=0,1,...$ are uniformly i.i.d. drawn from the training set $S=\{(x_i,y_i)\}_{i=1}^{| S|}$ where $S$ is an empirical version of the clean distribution $\mathcal{D}$. Then

(1) There exists $\mu>0$, Let $\mathcal{A}_1$: $\forall w_{\perp}\in\{w_{\perp}:\langle p,w_{\perp}\rangle=0\}$, $P_{(x,y)\sim S}\{y=1,f(x;w_{\perp})=0\}\ge\mu$, $P_{(x,y)\sim S}\{y=-1,f(x;w_{\perp})=0\}\ge\mu$. Then $P(\mathcal{A}_1)=1-\mathcal{O}(\frac{1}{|S|})$.

(2) For any $\epsilon>0$, Let $\mathcal{A}_2$: $\forall w_{\perp}\in\{w_{\perp}:\langle p,w_{\perp}\rangle=0,{\|w_{\perp}\|}_2=1\}$, $|E_{(x,y)\sim S}[yf(x;w_{\perp}]|\le\epsilon$. Then $P(\mathcal{A}_1)=1-\mathcal{O}(\frac{1}{|S|\epsilon^{d+1}})$.

(3.1) Let ${\theta}_t$ be the angle between $a_ip$ and $w_i^{(t)}$, if $\mathcal{A}_1$ and $\mathcal{A}_2$ hold, then $\forall K>1$, $P(\sin{\theta}_t<\frac{4K\eta}{\mu}\epsilon)\rightarrow\frac{1}{K}$ when $t\rightarrow\infty$.

(3.2) If $\mathcal{A}_1$ and $\mathcal{A}_2$ hold and $|\{i|a_i=1\}|=|\{i|a_i=-1\}|$, the classification accuracy over $\mathcal{D}$ will be 1 with probability at least $1-\mathcal{O}(\frac{\eta h}{\mu}\epsilon)$.
\end{theorem}

The theorem above shows that with $\epsilon>0$ small enough and $\mathcal{A}_1$, $\mathcal{A}_2$ hold, the filters can align the key pattern sufficiently well, and the network can achieve accuracy $1$ with probability close to $1$. And no matter how small $\epsilon$ is, $P(\mathcal{A}_1), P(\mathcal{A}_2)\rightarrow 1$ when $|S| \rightarrow \infty$.

%Combine Theorem~\ref{maxnoisy} and \ref{limiteddata}, we can provide a more general result considering empirical loss from noisy distribution, as Theorem~\ref{general} shows.

%\begin{theorem}\label{general}
%(\textbf{Empirical Loss from Noisy Data Distribution}) 
%Assumed that $p_+=-p_-=p$, $(x^{(t+1)},y^{(t+1)}),t=0,1,...$ are uniformly i.i.d. drawn from the training set $S_{\epsilon}=\{(x_i,y_i)\}_{i=1}^{| S_{\epsilon}|}$ where $S_{\epsilon}$ is an empirical version of the noisy distribution $\mathcal{D_{\epsilon}}$, where $\epsilon$ satisfies $0<\epsilon<2^{-2k-1}$. Then

%(1) Let ${\theta}_t$ be the angle between $a_ip$ and $w_i^{(t)}$, then with any large probability, if $|S_{\epsilon}|$ is sufficiently large, then $\sin{\theta}_t$ can be any small after enough steps;

%(2) With any large probability, if $|S_{\epsilon}|$ is sufficiently large, then after enough steps, the classification accuracy over $\Dcal_{\epsilon}$ will be 1.
%\end{theorem}

%The results of Theorem~\ref{limiteddata} and \ref{general} show that the learning of the CNN with SGD can generalize very well although training with limited data.

\subsection{Experiments}

To verify the theoretical results, we  conduct two classes of experiments. The first one is on the synthetic data, but the setting is more general than that stated in our theorems.  Second class of experiments is on the MNIST data to empirically demonstrate that our findings also hold for practical data though the assumptions cannot be strictly satisfied. 

\subsubsection{Synthetic Data}
The data distribution we used in this experiment is defined as follows.
\begin{definition}
(The General $\epsilon$-Noisy Distribution $\mathcal{D}^{\epsilon}$): Let $p_+,p_-\in\{p\in R^d:{||p||}_2=1\}$ be the positive  and negative key pattern ($p_+\ne p_-$), and let $(x,y)$ be a sample drawn from $\mathcal{D}^{\epsilon}$, where $x\in R^D$, $y\in\{-1,1\}$, then $x$ can be decomposed by $ x=x_0+x_1,$  
where
(1) $(x_0,y)$ is a sample draw from the clean distribution $\mathcal{D}$; 
(2) For any $u\in[k]$, $x_{1,ud+1...(u+1)d}$ is an i.i.d. sample drawn from $U(\{p:{\|p\|}_2\le\epsilon\})$.
\end{definition}
\emph{Remark}: The main difference between $\mathcal{D}_{\epsilon}$ and $\mathcal{D}^{\epsilon}$ is that, there is no noise in key patterns of data drawn from $\mathcal{D}_{\epsilon}$, but for data drawn from $\mathcal{D}^{\epsilon}$ there may be some noise in key patterns.

We let $D=100$, $d=10$, $k=10$, $\epsilon={10}^{-3}$, $p_+$ and $p_-$ be generated by standard normal distribution, then we obtain a  training set $S_{train}$ and testing set $S_{test}$ by i.i.d. sampling from $\mathcal{D}^{\epsilon}$ and $|S_{train}|=|S_{test}|=1000$. 
We set the number of filters $h=50$, the total number of iteration $10^4$, learning rate $10^{-2}$, batch size one in the learning process.

\begin{figure}[ht]
\vspace{-7mm}
\centering
\subfloat[\label{exp1.1.a}]{\includegraphics[width=0.33\textwidth]{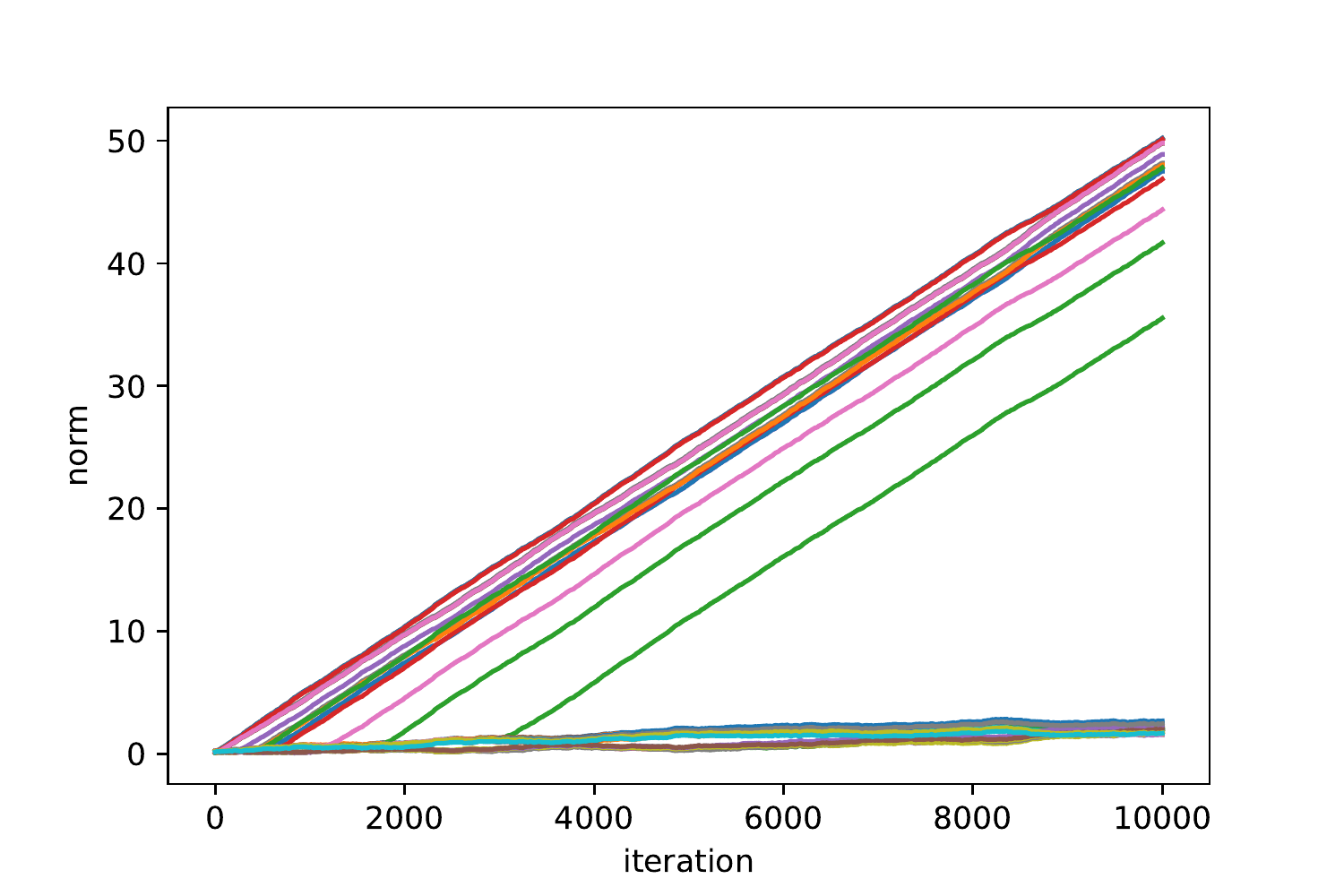}}\hfill
\subfloat[\label{exp1.1.b}]{\includegraphics[width=0.33\textwidth]{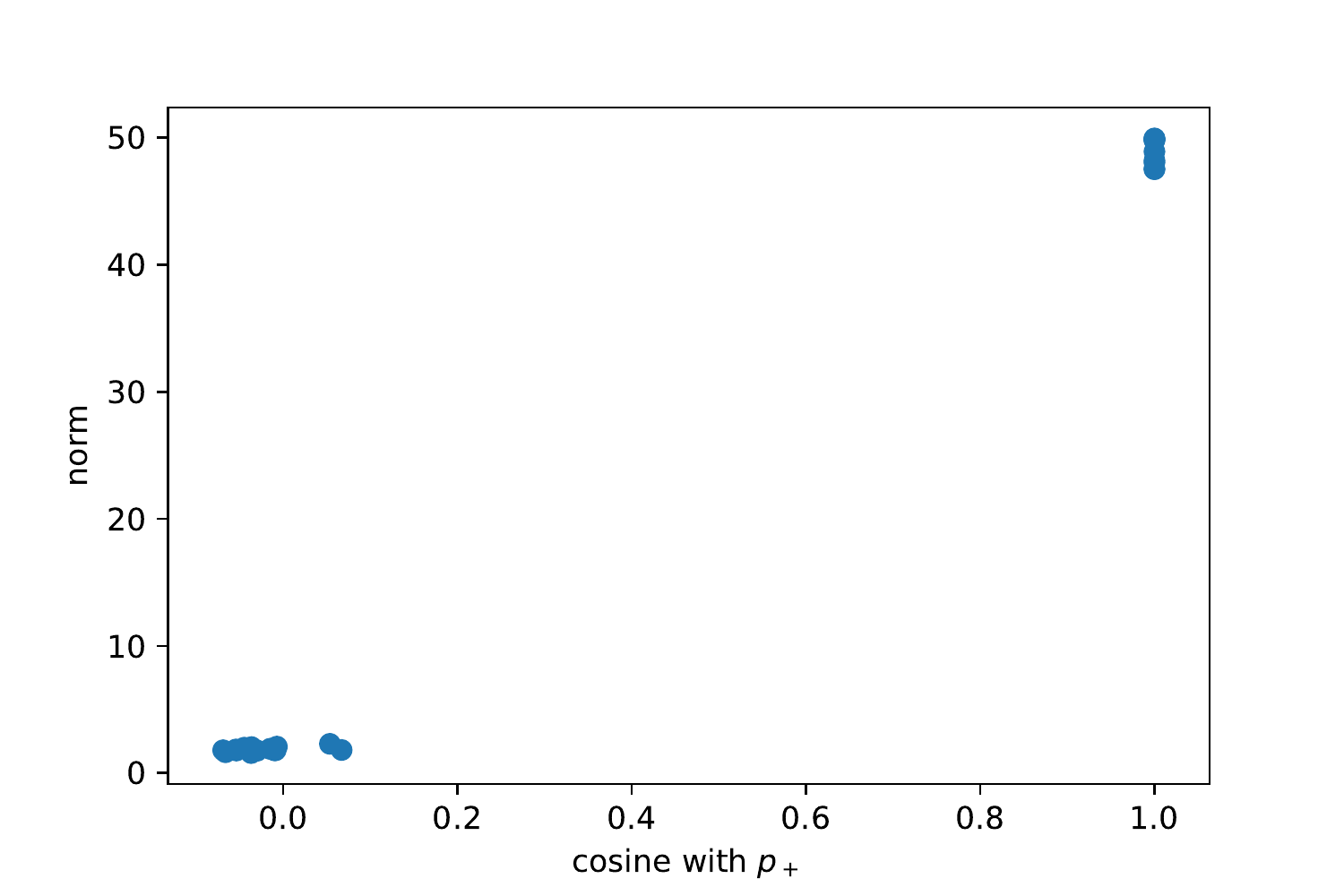}}\hfill
\subfloat[\label{exp1.1.c}]{\includegraphics[width=0.33\textwidth]{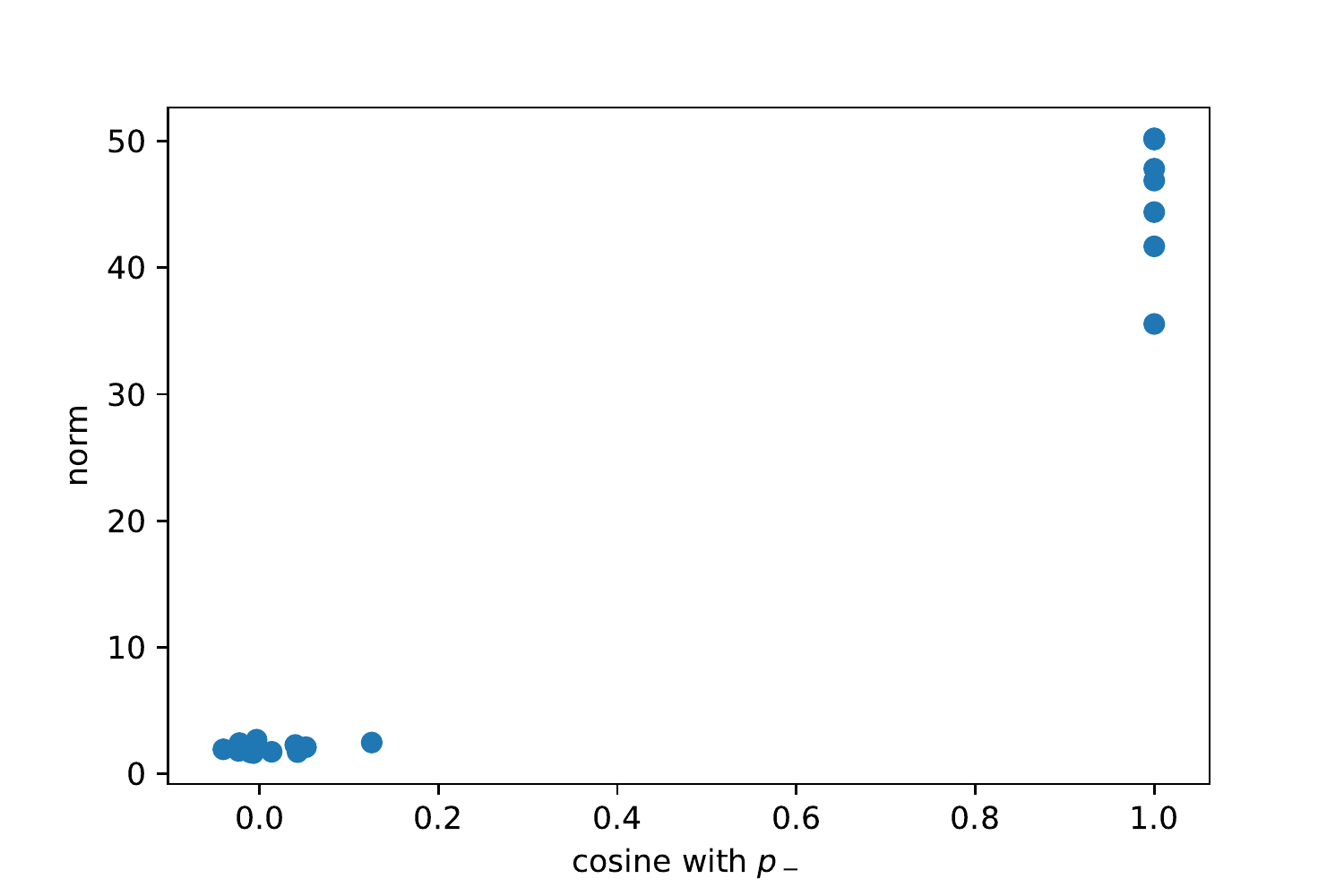}}
\vspace{-2mm}
\caption{Results on the synthetic data with an obtuse angle, $\langle p_+,p_-\rangle=-0.321$. (\textbf{a}) iteration v.s. $\|w_i\|_2$  for $50$ filters. (\textbf{b}) $\cos \angle (w_i,p_+)$ v.s. $\|w_i\|_2$ when $a_i=1$. (\textbf{c}) $\cos \angle (w_i,p_-)$ v.s. $\|w_i\|_2$ when $a_i=-1$.} \label{exp1.1}
\end{figure}

Figure~\ref{exp1.1} shows the results on the generated synthetic data when the positive and negative key pattern has an obtuse angle, $\langle p_+,p_-\rangle=-0.321$.
We can easily observe that: (1) the norm of some filters increases much faster than that of others; (2) when $a_i=1$, the norm of filters that have similar direction to $p_+$ is much larger; and when $a_i=-1$, the norm of filters with similar direction to $p_-$ is much larger.

\begin{figure}[ht]
\vskip -7mm
\centering
\subfloat[\label{exp1.2.a}]{\includegraphics[width=0.33\textwidth]{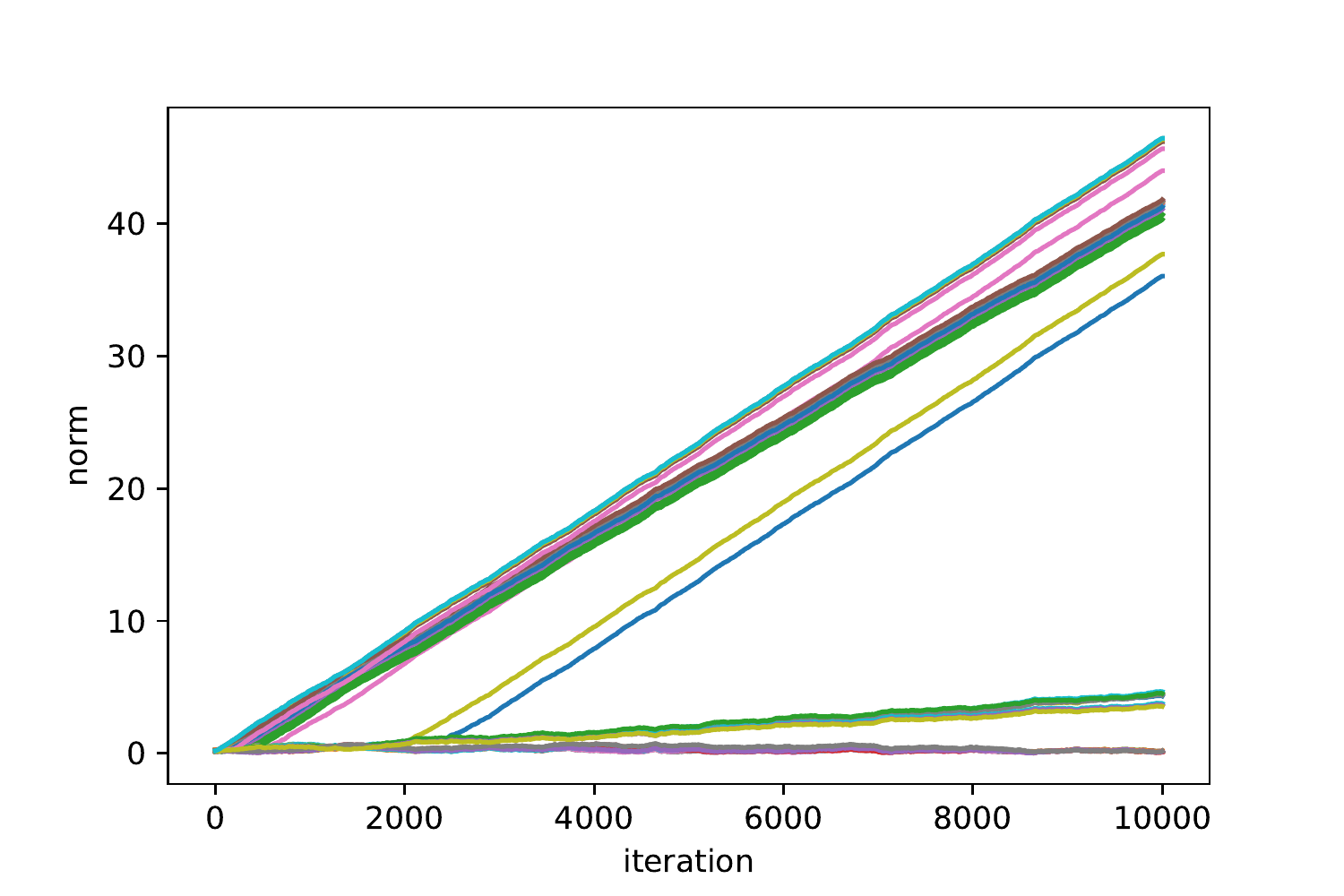}}\hfill
\subfloat[\label{exp1.2.b}]{\includegraphics[width=0.33\textwidth]{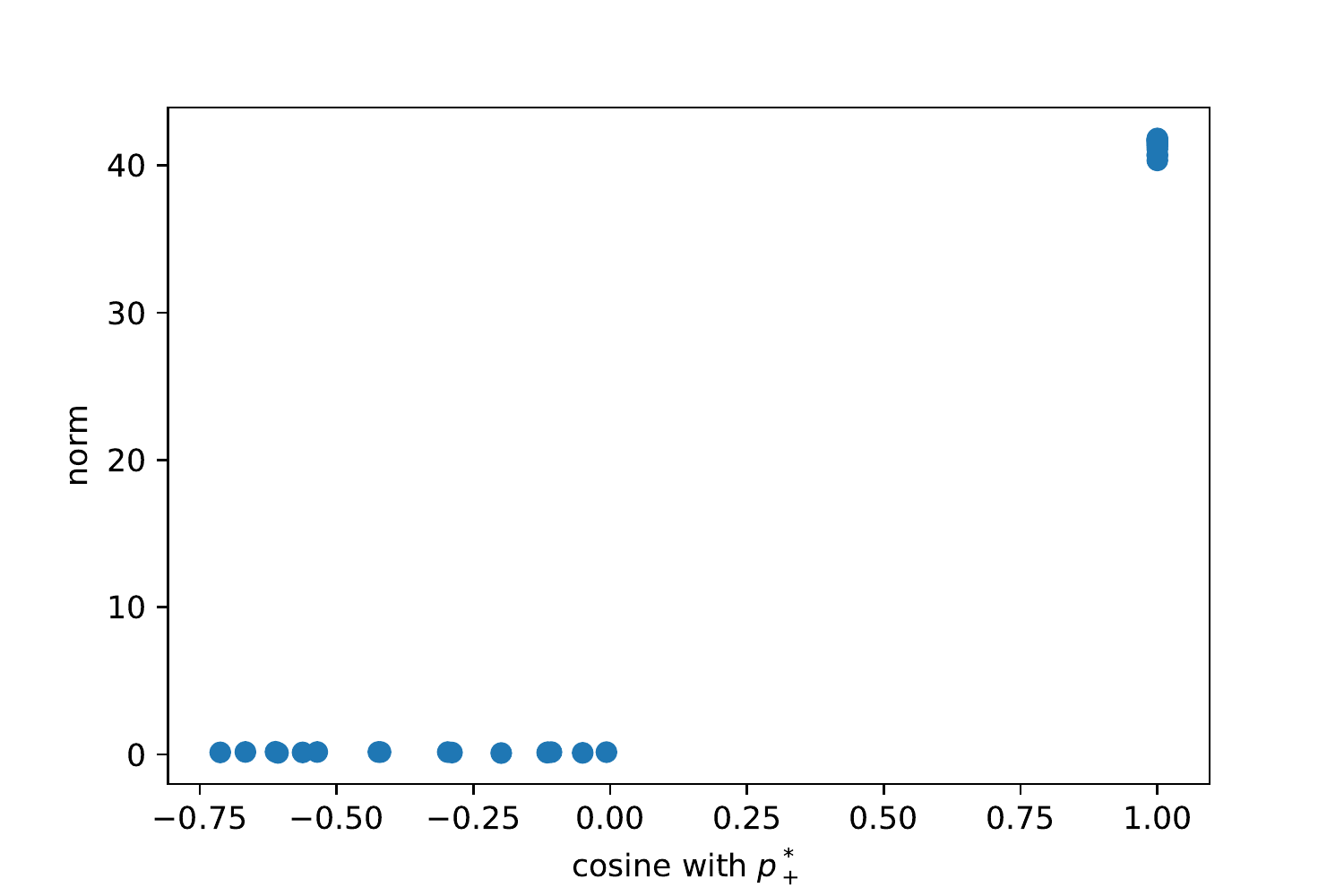}}\hfill
\subfloat[\label{exp1.2.c}]{\includegraphics[width=0.33\textwidth]{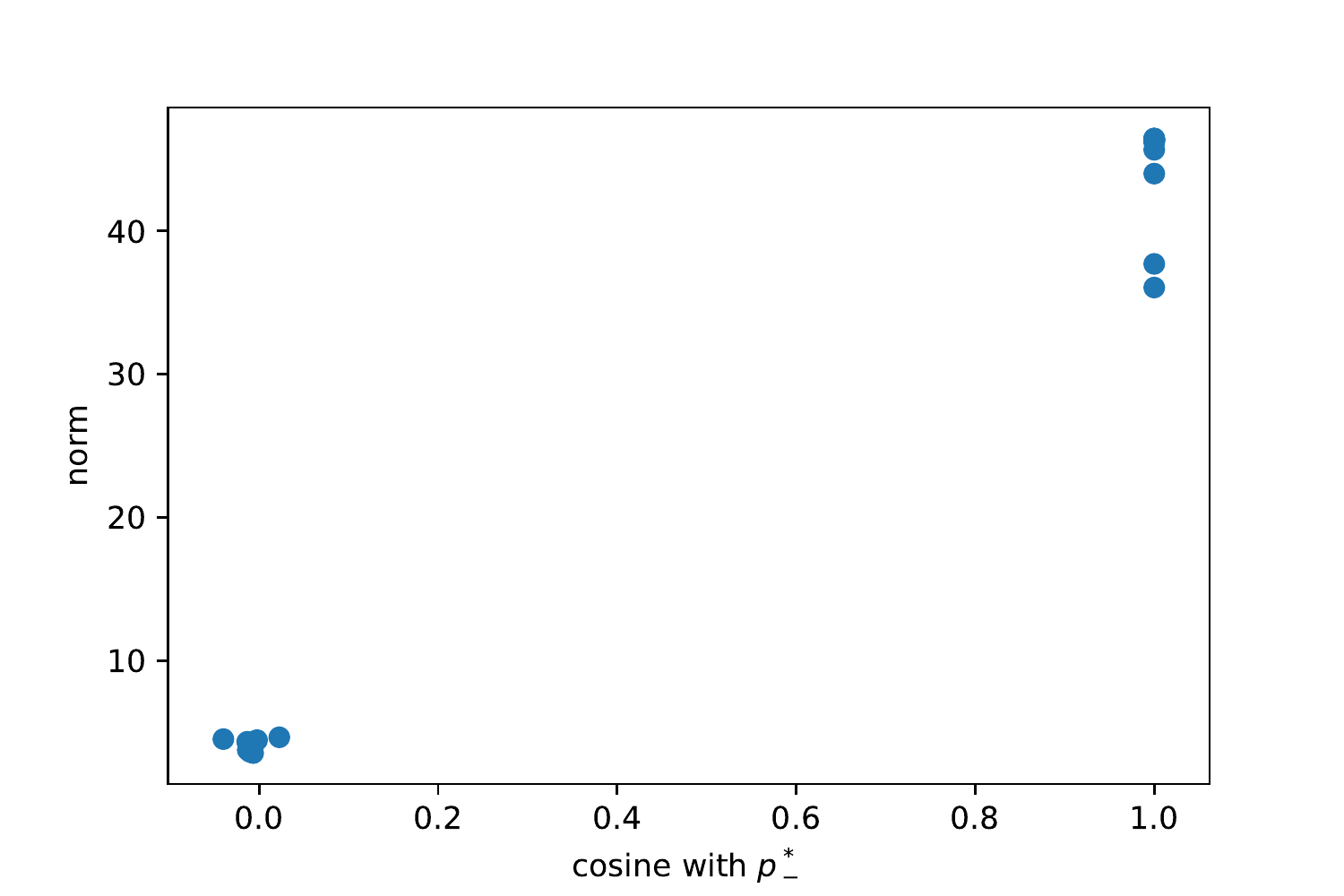}}
\vspace{-2mm}
\caption{Results on the synthetic data with an sharp angle, $\langle p_+,p_-\rangle=0.472$. (\textbf{a}) iteration v.s. $\| w_i \|_2$ for $50$ filters. (\textbf{b}) $\cos \angle (w_i,p_+^*)$ v.s. $\|w_i\|_2$ when $a_i=1$. (\textbf{c}) $\cos \angle (w_i,p_-^*)$ v.s. $\|w_i\|_2$ when $a_i=-1$.} \label{exp1.2}
\end{figure}

Figure~\ref{exp1.2} shows the results with an sharp angle, $\langle p_+,p_-\rangle=0.472$.
It can be seen that the norm of some filters increases much faster than that of others; and when $a_i=1$, the norm of filters that have similar direction to $p_+^*$ is much larger, where $p_+^*=p_+-\langle p_+,p_-\rangle p_-$; and when $a_i=-1$, the norm of filters with similar direction to $p_-^*$ is much larger, where $p_-^*=p_--\langle p_+,p_-\rangle p_+$.

\subsubsection{Variants of MNIST Data\label{subsection:mnist_nine}}
\begin{figure}[ht]
\vspace{-6mm}
\subfloat[\label{mnist_example}]{\includegraphics[scale=0.32]{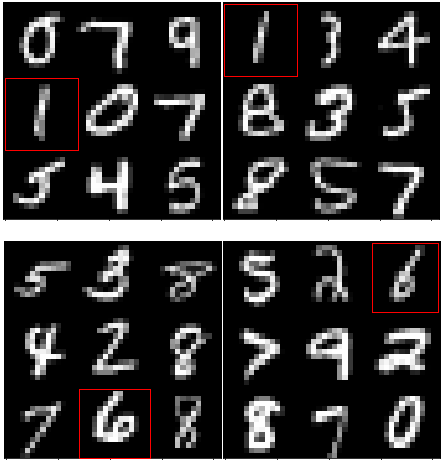}}\qquad\quad
\subfloat[\label{mnist_ker}]{\includegraphics[scale=0.8]{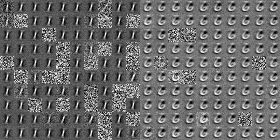}}
\vspace{-2mm}
\caption{(\textbf{a}) Top: two positive examples with the pattern ``1''. Bottom: two negative examples with the pattern ``6''. (\textbf{b}) Left: filters $w_i$ when $a_i=1$. Right: filters $w_i$ when $a_i=-1$.}
\end{figure}

In this part, we attempt to check whether our theoretical findings under strict assumptions still hold in a practical image classification task.  
We construct a new dataset that each image in this dataset consists of nine $14\times 14$ subsampled MNIST images. For positive examples, there is a number ``1'' and other eight positions are random numbers. For negative examples, there is a number ``6'' and others are random numbers. See Figure~\ref{mnist_example} for illustration. 
%Note that each image in the training set is composed of the subsampled images from the original MNIST training set, while each in the test set is composed of small images from the original MNIST test set.

We train a two-layer convolution neural network, the first layer is composed of $200$ filters with $14\times 14$ with stride size =[14,14], then the output is fed into a ReLU and global max-pooling layer. The second layer is a $200\times 1$ fully connected layer with fixed weight as $\pm 1$.  
% SGD is used and the batch size is $1$.

\textbf{Results.}  Figure \ref{mnist_ker} visualizes the 200 learned filters. We can easily observe that  most of the filters $w_i$ corresponding to $a_i=1$ successfully capture the pattern ``1'', and most of the filters $w_i$ corresponding to $a_i=-1$ successfully grab pattern ``6''. The empirical observations verify our theoretical results on the behavior of the convolutional filters.

\vspace{-5mm}
\section{Extensions: Multiple Key Patterns and Joint Training over All Parameters}\label{extent}

In this section, we extend our analysis on the learning dynamics of filters to two new scenarios. The first one is to consider that each data point have multiple key patterns determining its label. And the second one is to characterize joint learning of filters and output weights. 

\subsection{Learning with Multiple Key Patterns}
Real data often have more than one key pattern in an image (take the eyes and legs of a dog for an example), or the size of filter is smaller than that of the key pattern. 
%This scenario is more practical since typical CNNs only employs small filters like $3\times3$ or $5\times 5$.  
And for some special  case of multiple key patterns, we still can prove similar results in Section~\ref{mainresults}, i.e. different filters will capture different key patterns or the different parts of the large key pattern. In order to give a example, we define a multi-key pattern data distribution in Definition~\ref{multidata} (i.e. two key patterns) and present our theoretical results in Theorem~\ref{multipattern}.

\begin{definition}\label{multidata}
(\textbf{The Multi-Pattern Clean Distribution} $\mathcal{D}_{mul}$):
Let $p_{+,1},p_{+,2}\in\{p\in\Rbb ^d:{||p||}_2=1\}$ be the positive key patterns and $p_{-,1},p_{-,2}\in\{p\in\Rbb ^d:{||p||}_2=1\}$ be the negative patterns, and let $(x,y)$ be a sample drawn from $\mathcal{D}$, where $x\in\Rbb^D$, $y\in\{-1,1\}$, then

(1) $P\{y=1\}=P\{y=-1\}=\frac{1}{2}$;

(2) If $y=1$, there exists $u_1^*,u_2^*\in[k]$, s.t. $x_{u_1^*d+1,\dots,(u_2^*+1)d}=p_{+,1}$, $x_{u_2^*d+1,\dots,(u_2^*+1)d}=p_{+,2}$, and for any $u\in[k]$, $u\ne u_1^*,u_2^*$, $x_{ud+1,\dots,(u+1)d}$ is an i.i.d. sample draw from $U(\{p\in \Rbb^d:p\perp span\{p_{+,1},p_{+,2},p_{-,1},p_{-,2}\},{||p||}_2=1\})$, where $U(S)$ is the uniform distribution over set $S$;

(3) If $y=-1$, there exists $u_1^*,u_2^*\in[k]$, s.t. $x_{u_1^*d+1,\dots,(u_2^*+1)d}=p_{-,1}$, $x_{u_2^*d+1,\dots,(u_2^*+1)d}=p_{-,2}$, and for any $u\in[k]$, $u\ne u_1^*,u_2^*$, $x_{ud+1,\dots,(u+1)d}$ is an i.i.d. sample draw from $U(\{p\in \Rbb^d:p\perp span\{p_{+,1},p_{+,2},p_{-,1},p_{-,2}\},{||p||}_2=1\})$, where $U(S)$ is the uniform distribution over set $S$.
\end{definition}

\begin{theorem}\label{multipattern}
(\textbf{Learning with Multiple Key Patterns}) 
Assumed that $(x^{(t+1)},y^{(t+1)}),t=0,1,...$ are i.i.d. drawn from the multi pattern clean distribution $\mathcal{D}_{mul}$ and $p_{+,1}$, $p_{+,2}$, $p_{-,1}$, $p_{-,2}$ are perpendicular to each other. Then

(1) When $a_i=1$, if $\langle p_{+,1},w_i^{(0)}\rangle>\langle p_{+,2},w_i^{(0)}\rangle$, let ${\theta}_t$ be the angle between $p_{+,1}$ and $w_i^{(t)}$, if $\langle p_{+,2},w_i^{(0)}\rangle>\langle p_{+,1},w_i^{(0)}\rangle$, let ${\theta}_t$ be the angle between $p_{+,2}$ and $w_i^{(t)}$. If $max\{\langle p_{+,1},w_i^{(0)}\rangle,\langle p_{+,2},w_i^{(0)}\rangle\}>0$ hold, then

\begin{equation}
P\{\sin{\theta}_t\le \mathcal{O}(t^{-(\frac{1}{2}-\sigma)})\}\ge1-\mathcal{O}(t^{-\sigma})
\end{equation}

(2) When $a_i=-1$, if $\langle p_{-,1},w_i^{(0)}\rangle>\langle p_{-,2},w_i^{(0)}\rangle$, let ${\theta}_t$ be the angle between $p_{-,1}$ and $w_i^{(t)}$, if $\langle p_{-,2},w_i^{(0)}\rangle>\langle p_{-,1},w_i^{(0)}\rangle$, let ${\theta}_t$ be the angle between $p_{-,2}$ and $w_i^{(t)}$. If $max\{\langle p_{-,1},w_i^{(0)}\rangle,\langle p_{-,2},w_i^{(0)}\rangle\}>0$ hold, then

\begin{equation}
P\{\sin{\theta}_t\le \mathcal{O}(t^{-(\frac{1}{2}-\sigma)})\}\ge1-\mathcal{O}(t^{-\sigma})
\end{equation}
\end{theorem}

\begin{figure}[ht]
\vspace{-6mm}
\subfloat[\label{mnist_example2}]{\includegraphics[scale=0.32]{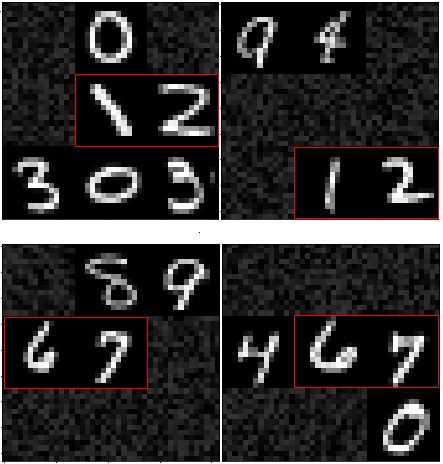}}\qquad\quad
\subfloat[\label{mnist_ker2}]{\includegraphics[scale=0.8]{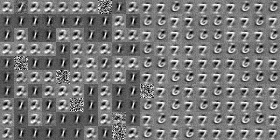}}
\vspace{-2mm}
\caption{(\textbf{a}) Top: two positive examples with the pattern ``12''. Bottom: two negative examples with the pattern ``67''. (\textbf{b}) Left: filters $w_i$ when $a_i=1$. Right: filters $w_i$ when $a_i=-1$.}
\end{figure}

To empirically verify this multi-pattern case, we construct the following dataset. For positive images, there is the pattern ``12'' and other slots are random numbers or noise. For negative examples, there exists the pattern ``67'' and others  are random numbers or noises, as shown in Figure \ref{mnist_example2}. The network architecture and settings are used as in section \ref{subsection:mnist_nine}. It can be observed from Figure~\ref{mnist_ker2} that different filters could learn different parts of the key pattern.

\vspace{-4mm}
\subsection{Jointly Training $a_i$ and $w_i$}
The theorems above do not analyze the situation that $a_i$ is also trained. If the weight of full-connected layer $a_i$ and filters $w_i$ are jointly trained, we will have the following conclusion.

\begin{theorem}\label{cotrain}
(\textbf{Joint Training})  
Assumed that every mini-batch is i.i.d. randomly drawn from the clean distribution $\mathcal{D}$ with as many positive examples as negative examples, and $p_+=-p_-=p$, with initialization $a_i^{(0)}=\pm 1$, and $||w_i^{(0)}||<\eta \ll 1$, let $w_i^{(t)}=\alpha_i^{(t)}p+w_{i,\perp}^{(t)}$, then $sgn(a_i^{(t)})=sgn(a_i^{(0)})$, $a_i^{(t)}\rightarrow\infty$ and for $t>\frac{1}{\eta}$, $\alpha_i^{(t)}=sgn(a_i^{(0)})$,$\alpha_i^{(t)}>||w_{i,\perp}^{(t)}||$, hold with probability $p=1-\mathcal{O}(\eta)$.
\end{theorem}

This theorem shows that if the filters and learning rate $\eta$ is initialized sufficiently small, then with a high probability, $sgn(a_i^{(t)})$ will not change, and $w_{i,\parallel}^{(t)}$ is always dominant during joint training.

\section{Conclusion \& Future Work}

In this work we analyze the learning dynamics  of a two-layer nonlinear CNN with stochastic gradient descent (SGD) for binary classification task. We prove that when training the CNN over some relatively realistic data distributions (e.g. image-like data distribution), the convolutional filters will learn the key patterns from  the data and the norm of these filters will increase much faster than that of others. And then with large probability, the classification accuracy of the learned CNN will be 1 over the considered distribution. 

We believe that the bias introduced by the training  of the two-layer CNN is important for understanding why CNN works.  Moreover, it is also crucial to find whether there is a similar bias in  deep CNNs. We leave the investigation along this line as future work. And we think that this bias also can help us to develop the improved training algorithm for CNNs or conduct channel pruning.

\bibliographystyle{plain}
\bibliography{cnn_learning}

\section*{Appendix}

\subsection*{Notations}

We use $\mathcal{F}_t$ to stand for training data sequence until steps $t$: $\{(x^{(0)},y^{(0)}),\dots,(x^{(t)},y^{(t)})\}$, use $[k]$ to stand for $1,\dots,k$, use $sgn$ to stand for sign function. And we say ``$x_{ud+1...(u+1)d}^{(t+1)}$ is activated by $w_i^{(t)}$'' or ``$x_{ud+1...(u+1)d}^{(t+1)}$ is activated'' if $u={\argmax}_v\{\langle x_{vd+1...(v+1)d}^{(t+1)},w_i^{(t)}\rangle\}$ in equation (7).

\subsection*{Proofs for Section 2}

\paragraph{Proposition 1}
Assumed that $p_+=-p_-=p$, then when $k\ge\frac{2}{\epsilon}+1$, the $\epsilon$-noisy distribution ${\Dcal}_{\epsilon}$ is not linearly separable.

\paragraph{Proof}Assume that there exist $w\in\Rbb^D$ and $b\in\Rbb$ s.t. $\forall (x,y)\sim{\Dcal}_{\epsilon}$, $y(w^Tx+b)>0$, then we have that

\begin{eqnarray*}
w^T(p;0;\dots;0)+b&>&0\\
w^T(-p;0;\dots;0)+b&<&0\\
w^T(0;p;\dots;0)+b&>&0\\
w^T(0;-p;\dots;0)+b&<&0\\
&\dots&\\
w^T(0;0;\dots;p)+b&>&0\\
w^T(0;0;\dots;-p)+b&<&0\\
\end{eqnarray*}

So we have that $-w_{ud+1\dots(u+1)d}^Tp<b<w_{ud+1\dots(u+1)d}^Tp$, $\forall u$.

Without loss of generality, we assume that

\begin{eqnarray*}
w_{1\dots d}^Tp=min_{u}\{w_{ud+1\dots(u+1)d}^Tp\}
\end{eqnarray*}

Then Let $x=(p;-\epsilon p;\dots;-\epsilon p),y=1$ be a sample drawn from ${\Dcal}_{\epsilon}$, then we have that

\begin{eqnarray*}
w^Tx+b&=&w_{1\dots d}^Tp+b-\epsilon\sum_{u>1}w_{ud+1\dots(u+1)d}^Tp\\
&\le&(1-\epsilon(k-1))w_{1\dots d}^Tp+b\\
&\le&-w_{1\dots d}^Tp+b\\
&<&0
\end{eqnarray*}

Which leads to a contradiction.

\paragraph{Proposition 2}
\textbf{(Non-convex and Non-concave Examples)}
Let $D=4$, $d=2$, $p_+={(1,0)}^T$, $p_-={(-1,0)}^T$, $h=1$, $a_1=-1$, then the optimization problem 
$
{\min}_{W}E_{(x,y)\sim\Dcal}[\ell(F(x;W),y)]
$ 
is not convex and not concave.

\paragraph{Proof}Let $w_1={(w_{11},w_{12})}^T$ and let $\phi(w_{11},w_{12})=E_{(x,y)\sim\Dcal}[\ell(F(x;W),y)]$. Note that $\{w\in \Rbb^d:\langle p_+,w\rangle=0,\langle p_-,w\rangle=0,{\|w\|}_2=1\}=\{{(0,1)}^T,{(0,-1)}^T\}$, then

\begin{eqnarray*}
\phi(w_{11},w_{12})&=&E_{(x,y)\sim\Dcal}[\ell(F(x;W),y)]\\
&=&E(ymax\{0,\langle x_{12},w_1\rangle,\langle x_{34},w_1\rangle\})\\
&=&\frac{1}{4}(max\{0,w_{11},w_{12}\}+max\{0,w_{11},-w_{12}\}\\
&-&max\{0,-w_{11},w_{12}\}-max\{0,-w_{11},-w_{12}\})
\end{eqnarray*}

Let $w^{(1)}={(4,4)}^T$, $w^{(2)}={(3,-1)}^T$, then we have that

\begin{eqnarray*}
\phi(w^{(1)})+\phi(w^{(2)})-2\phi(\frac{w^{(1)}+w^{(2)}}{2})&=&1+1.25-2\times1.375\\
&=&-0.5<0
\end{eqnarray*}

Let $w^{(1)}={(4,-3)}^T$, $w^{(2)}={(-3,-3)}^T$, then we have that

\begin{eqnarray*}
\phi(w^{(1)})+\phi(w^{(2)})-2\phi(\frac{w^{(1)}+w^{(2)}}{2})&=&1.25-0.75-2\times0.125\\
&=&0.25>0
\end{eqnarray*}

So $E_{(x,y)\sim\Dcal}[\ell(F(x;W),y)]$ is not convex and not concave.

\subsection*{Proofs for Section 3}

\paragraph{Theorem 1}
(\textbf{Learning with Clean Data Distribution}) 
Assumed that $(x^{(t+1)},y^{(t+1)}),t=0,1,\dots$ are i.i.d. randomly drawn from the clean distribution $\Dcal$. Then

(1) When $p_+=-p_-=p$,

(1.1) For any $\sigma\in(0,\frac{1}{2})$, let ${\theta}_t$ be the angle between $a_ip$ and $w_i^{(t)}$, then
\begin{equation*}
   P\{\sin{\theta}_t\le O(t^{-(\frac{1}{2}-\sigma)})\}\ge1-O(t^{-\sigma}) 
\end{equation*}

(1.2) For any $\sigma\in(0,\frac{1}{2})$, $\epsilon>0$, when
\begin{equation*}
t\ge max\{\frac{(2^{k+2}-4)(h+2)}{\epsilon},{(\frac{(\eta+2)(h+2)}{\epsilon})}^{\frac{1}{\sigma}},{(\frac{2^{k+1}(2h+1)}{\eta})}^{\frac{2}{1-2\sigma}}\}
\end{equation*}
then with probability at least $1-{(\frac{1}{2})}^{h-1}-\epsilon$, after $t$ steps, the classification accuracy over $\Dcal$ will be $1$;

(2) When $\langle p_+,p_-\rangle\le0$ and $p_+\ne-p_-$,

(2.1) If $a_i=1$, let ${\theta}_t$ be the angle between $p_+$ and $w_i^{(t)}$, otherwise let ${\theta}_t$ be the angle between $p_-$ and $w_i^{(t)}$.
Decomposed $w_i^{(t)}$ by $w_i^{(t)}=w_{\parallel,i}^{(t)}+w_{\perp,i}^{(t)}$, where $w_{\parallel,i}^{(t)}\in span\{p_+,p_-\}$, $w_{\perp,i}^{(t)}$ is perpenticular to $p_+$ and $p_-$.
If $w_{\parallel,i}^{(0)}$ is in the shaded region of Figure 1(a), then for any $\sigma\in(0,\frac{1}{2})$,
\begin{equation*}
   P\{\sin{\theta}_t\le O(t^{-(\frac{1}{2}-\sigma)})\}\ge1-O(t^{-\sigma}) 
\end{equation*}

(2.2) For any $\sigma\in(0,\frac{1}{2}),\epsilon>0$, when
\begin{equation*}
t\ge max\{\frac{(2^{k+2}-4)(h+2)}{\epsilon},{(\frac{(\eta+2)(h+2)}{\epsilon})}^{\frac{1}{\sigma}},{(\frac{2^{k+1}(2h+1)}{\eta})}^{\frac{2}{1-2\sigma}}\}
\end{equation*}
then with probability at least $1-2\cdot{(\frac{7}{8})}^h-\epsilon$, after $t$ steps the classification accuracy over $\Dcal$ will be $1$;

(3) When $\langle p_+,p_-\rangle>0$, $p_+\ne p_-$, positive and negative examples sample alternatively,

(3.1) Let $T$ be a positive integer. If $a_i=1$, let ${\theta}_t$ be the angle between $p_+-\langle p_+,p_-\rangle p_-$ and $w_i^{(T+t)}$, otherwise let ${\theta}_t$ be the angle between $p_--\langle p_+,p_-\rangle p_+$ and $w_i^{(T+t)}$.
Decomposed $w_i^{(t)}$ by $w_i^{(t)}=w_{\parallel,i}^{(t)}+w_{\perp,i}^{(t)}$, where $w_{\parallel,i}^{(t)}\in span\{p_+,p_-\}$, $w_{\perp,i}^{(t)}$ is perpenticular to $p_+$ and $p_-$.
If $T$ is sufficiently large and $w_{\parallel,i}^{(T)}$ is in the shaded region of Figure 1(b), then for any $\sigma\in(0,\frac{1}{2})$,
\begin{equation*}
P\{\sin{\theta}_{t+T}\le O(t^{-(\frac{1}{2}-\sigma)})\}\ge1-O(t^{-\sigma})-O(T^{-\sigma})
\end{equation*}

(3.2) For any $\sigma\in(0,\frac{1}{2}),\epsilon>0$, when $T$ and $t$ are sufficiently large, then with probability at least $1-2\cdot{(\frac{7}{8})}^h-\epsilon$, after $T+t$ steps the classification accuracy over $\Dcal$ will be $1$.

\begin{figure}[ht]
    \centering
    \subfloat[\label{dunnjiao}]{\includegraphics[width=0.35\textwidth]{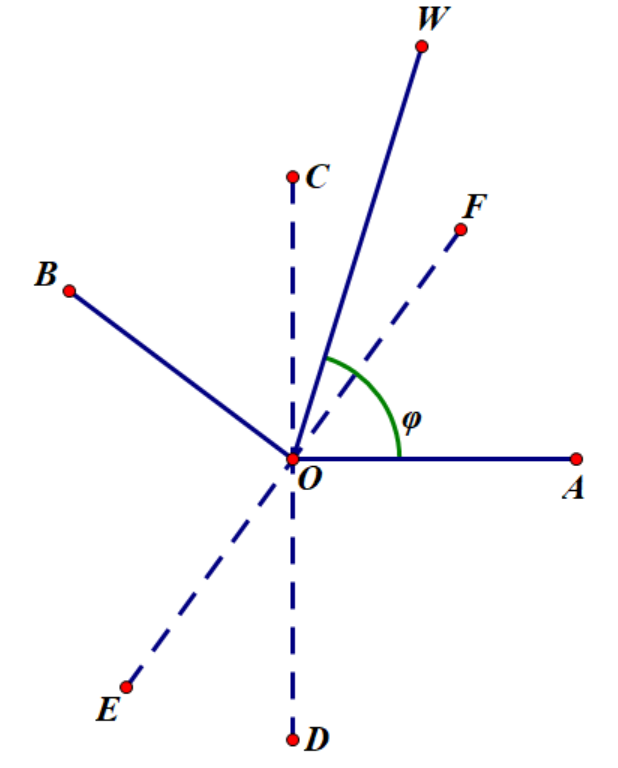}}\hfill
    \subfloat[\label{ruijiao}]{\includegraphics[width=0.35\textwidth]{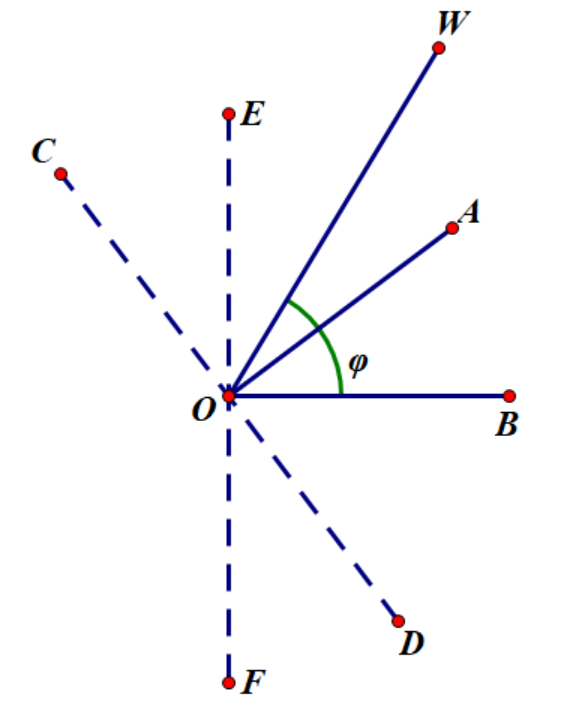}}\hfill
    \vspace{-2mm}
    \caption{Illustration for proof of Theorem 1. (a) For proof of Theorem 1(2). (b) For proof of Theorem 1(3).}
    \label{dunjiaoruijiao}
\end{figure}

\paragraph{Proof}Without loss of generality, we assume that $u^*$ in Definition 1 is equaled to $1$.

(1)When $p_+=-p_-=p$,

Consider the decomposition of $w_i^{(t)}$

\begin{eqnarray*}
w_i^{(t)}={\alpha}_i^{(t)}p+w_{\perp,i}^{(t)}
\end{eqnarray*}

where ${\alpha}_i^{(t)}\in\Rbb$ and $w_{\perp,i}^{(t)}$ is perpendicular to $p$.

(a) First we will analyze the property of ${\alpha}_i^{(t)}$.

Without loss of generality, we assume that ${\alpha}_i^{(t)}\ne0$.

(a.1)When $a_i=1$,

According to the dynamics of $w_i^{(t)}$, we have that

\begin{eqnarray*}
{\alpha}_i^{(t+1)}p&=&{\alpha}_i^{(t)}p+\eta y^{(t+1)}x_{1\dots d}^{(t+1)}\Ibb_{\{x_{1\dots d}^{(t+1)}\ is\ activated\}}\\
&=&{\alpha}_i^{(t)}p+\eta p\Ibb_{\{x_{1\dots d}^{(t+1)}\ is\ activated\}}
\end{eqnarray*}

Note that when $y^{(t+1)}=sgn({\alpha}_i^{(t)})$ and $f(x^{(t+1)};w_{\perp,i}^{(t)})=0$, $x_{1\dots d}^{(t+1)}$ must be activated by $w_i^{(t)}$, so we have that

\begin{eqnarray*}
{\alpha}_i^{(t+1)}&=&{\alpha}_i^{(t)}+\eta\Ibb_{\{x_{1\dots d}^{(t+1)}\ is\ activated\}}\\
&\ge&{\alpha}_i^{(t)}+\eta\Ibb_{\{y^{(t+1)}=sgn({\alpha}_i^{(t)}),f(x^{(t+1)};w_{\perp,i}^{(t)})=0\}}
\end{eqnarray*}

According to the definition of $\Dcal$, we have that

\begin{eqnarray*}
&&E[\Ibb_{\{y^{(t+1)}=sgn({\alpha}_i^{(t)}),f(x^{(t+1)};w_{\perp,i}^{(t)})=0\}}|\mathcal{F}_t]\\
&=&E[\Ibb_{\{y^{(t+1)}=sgn({\alpha}_i^{(t)}),\langle x_{d+1\dots 2d}^{(t+1)},w_{\perp,i}^{(t)}\rangle\le 0,\dots,\langle x_{D-d+1\dots D}^{(t+1)},w_{\perp,i}^{(t)}\rangle\le 0\}}|\mathcal{F}_t]\\
&=&2^{-k}
\end{eqnarray*}

So no matter what ${\alpha}_i^{(t)}$ and $w_{\perp,i}^{(t)}$ are, ${\alpha}_i^{(t+1)}={\alpha}_i^{(t)}+\eta$ will hold with probability at least $2^{-k}$ and if ${\alpha}_i^{(t+1)}={\alpha}_i^{(t)}+\eta$ does not hold, then ${\alpha}_i^{(t+1)}={\alpha}_i^{(t)}$ will hold. So ${\alpha}_i^{(t)}$ is monotonically increasing and there exists a sequence of i.i.d. Bernoulli random variables $e_1,\dots,e_t$, satisfies $P\{e_s=1\}=2^{-k},s=1,\dots,t$ and

\begin{eqnarray*}
{\alpha}_i^{(t)}\ge{\alpha}_i^{(0)}+\eta\sum_{s=1}^te_s
\end{eqnarray*}

So we have that

\begin{eqnarray*}
P\{{\alpha}_i^{(t)}\le{\alpha}_i^{(0)}+2^{-k-1}\eta t\}&\le&P\{\sum_{s=1}^te_s\le 2^{-k-1}t\}\\
&\le&P\{|\sum_{s=1}^te_s-2^{-k}t|\ge 2^{-k-1}t\}\\
&\le&\frac{2^{-k}(1-2^{-k})t}{{(2^{-k-1}t)}^2}\\
&=&\frac{2^{k+2}-4}{t}
\end{eqnarray*}

The last inequality is guaranteed by Chebyshev's Inequality.

(a.2)When $a_i=-1$,

Similar to (a.1), we can proof that ${\alpha}_i^{(t)}$ is monotonically decreasing and

\begin{eqnarray*}
P\{{\alpha}_i^{(t)}\ge{\alpha}_i^{(0)}-2^{-k-1}\eta t\}\le\frac{2^{k+2}-4}{t}
\end{eqnarray*}

(b) Next we will analyze the property of $w_{\perp,i}^{(t)}$.

(b.1)When $a_i=1$,

If ${\alpha}_i^{(t)}>0$, then according to the dynamics of $w_i^{(t)}$, we have that

\begin{eqnarray*}
w_{\perp,i}^{(t+1)}&=&w_{\perp,i}^{(t)}+\eta x_{u_1d+1\dots(u_1+1)d}^{(t+1)}\Ibb_{\{y^{(t+1)}=1,f(x^{(t+1)};w_{\perp,i}^{(t)})>{\alpha}_i^{(t)}\}}\\
&-&\eta x_{u_2d+1\dots(u_2+1)d}^{(t+1)}\Ibb_{\{y^{(t+1)}=-1,f(x^{(t+1)};w_{\perp,i}^{(t)})>0\}}
\end{eqnarray*}

where $u_1,u_2\in[k]$ and $u_1,u_2\ne1$.

So we have that

\begin{eqnarray*}
&&E[{\|w_{\perp,i}^{(t+1)}\|}_2^2|\mathcal{F}_t]\\
&=&{\|w_{\perp,i}^{(t)}\|}_2^2+{\eta}^2E[\Ibb_{\{y^{(t+1)}=1,f(x^{(t+1)};w_{\perp,i}^{(t)})>{\alpha}_i^{(t)}\}}|\mathcal{F}_t]\\
&+&{\eta}^2E[\Ibb_{\{y^{(t+1)}=-1,f(x^{(t+1)};w_{\perp,i}^{(t)})>0\}}|\mathcal{F}_t]\\
&+&2\eta E[\langle x_{u_1d+1\dots(u_1+1)d}^{(t+1)},w_{\perp,i}^{(t)}\rangle \Ibb_{\{y^{(t+1)}=1,f(x^{(t+1)};w_{\perp,i}^{(t)})>{\alpha}_i^{(t)}\}}|\mathcal{F}_t]\\
&-&2\eta E[\langle x_{u_2d+1\dots(u_2+1)d}^{(t+1)},w_{\perp,i}^{(t)}\rangle \Ibb_{\{y^{(t+1)}=-1,f(x^{(t+1)};w_{\perp,i}^{(t)})>0\}}|\mathcal{F}_t]\\
&\le&{\|w_{\perp,i}^{(t)}\|}_2^2+{\eta}^2+2\eta E[f(x^{(t+1)};w_{\perp,i}^{(t)})\Ibb_{\{y^{(t+1)}=1,f(x^{(t+1)};w_{\perp,i}^{(t)})>{\alpha}_i^{(t)}\}}|\mathcal{F}_t]\\
&-&2\eta E[f(x^{(t+1)};w_{\perp,i}^{(t)})\Ibb_{\{y^{(t+1)}=-1,f(x^{(t+1)};w_{\perp,i}^{(t)})>0\}}|\mathcal{F}_t]\\
&\le&{\|w_{\perp,i}^{(t)}\|}_2^2+{\eta}^2+2\eta E[f(x^{(t+1)};w_{\perp,i}^{(t)})\Ibb_{\{y^{(t+1)}=1,f(x^{(t+1)};w_{\perp,i}^{(t)})>0\}}|\mathcal{F}_t]\\
&-&2\eta E[f(x^{(t+1)};w_{\perp,i}^{(t)})\Ibb_{\{y^{(t+1)}=-1,f(x^{(t+1)};w_{\perp,i}^{(t)})>0\}}|\mathcal{F}_t]\\
&=&{\|w_{\perp,i}^{(t)}\|}_2^2+{\eta}^2+2\eta E[y^{(t+1)}f(x^{(t+1)};w_{\perp,i}^{(t)})|\mathcal{F}_t]\\
&=&{\|w_{\perp,i}^{(t)}\|}_2^2+{\eta}^2
\end{eqnarray*}

The first equality is because that

\begin{eqnarray*}
\Ibb_{\{y^{(t+1)}=1,f(x^{(t+1)};w_{\perp,i}^{(t)})>{\alpha}_i^{(t)}\}}\Ibb_{\{y^{(t+1)}=-1,f(x^{(t+1)};w_{\perp,i}^{(t)})>0\}}=0
\end{eqnarray*}

The first inequality is because that

\begin{eqnarray*}
&&E[\Ibb_{\{y^{(t+1)}=1,f(x^{(t+1)};w_{\perp,i}^{(t)})>{\alpha}_i^{(t)}\}}+\Ibb_{\{y^{(t+1)}=-1,f(x^{(t+1)};w_{\perp,i}^{(t)})>0\}}|\mathcal{F}_t]\\
&\le&E[\Ibb_{\{y^{(t+1)}=1\}}+\Ibb_{\{y^{(t+1)}=-1\}}|\mathcal{F}_t]\\
&=&1
\end{eqnarray*}

and

\begin{eqnarray*}
&&\langle x_{u_1d+1\dots(u_1+1)d}^{(t+1)},w_{\perp,i}^{(t)}\rangle \Ibb_{\{y^{(t+1)}=1,f(x^{(t+1)};w_{\perp,i}^{(t)})>{\alpha}_i^{(t)}\}}\\
&=&f(x^{(t+1)};w_{\perp,i}^{(t)})\Ibb_{\{y^{(t+1)}=1,f(x^{(t+1)};w_{\perp,i}^{(t)})>{\alpha}_i^{(t)}\}}\\
&&\langle x_{u_2d+1\dots(u_2+1)d}^{(t+1)},w_{\perp,i}^{(t)}\rangle \Ibb_{\{y^{(t+1)}=-1,f(x^{(t+1)};w_{\perp,i}^{(t)})>0\}}\\
&=&f(x^{(t+1)};w_{\perp,i}^{(t)})\Ibb_{\{y^{(t+1)}=-1,f(x^{(t+1)};w_{\perp,i}^{(t)})>0\}}
\end{eqnarray*}

The last equality is because that

\begin{eqnarray*}
&&E[y^{(t+1)}f(x^{(t+1)};w_{\perp,i}^{(t)})|\mathcal{F}_t]\\
&=&\frac{1}{2}E[max\{\langle x_{d+1\dots 2d}^{(t+1)},w_{\perp,i}^{(t)}\rangle,\dots,\langle x_{D-d+1\dots D}^{(t+1)},w_{\perp,i}^{(t)}\rangle\}|\mathcal{F}_t,y^{(t+1)}=1]\\
&-&\frac{1}{2}E[max\{\langle x_{d+1\dots 2d}^{(t+1)},w_{\perp,i}^{(t)}\rangle,\dots,\langle x_{D-d+1\dots D}^{(t+1)},w_{\perp,i}^{(t)}\rangle\}|\mathcal{F}_t,y^{(t+1)}=-1]\\
&=&\frac{1}{2}E[max\{\langle x_{d+1\dots 2d}^{(t+1)},w_{\perp,i}^{(t)}\rangle,\dots,\langle x_{D-d+1\dots D}^{(t+1)},w_{\perp,i}^{(t)}\rangle\}|\mathcal{F}_t]\\
&-&\frac{1}{2}E[max\{\langle x_{d+1\dots 2d}^{(t+1)},w_{\perp,i}^{(t)}\rangle,\dots,\langle x_{D-d+1\dots D}^{(t+1)},w_{\perp,i}^{(t)}\rangle\}|\mathcal{F}_t]\\
&=&0
\end{eqnarray*}

If ${\alpha}_i^{(t)}<0$, then we have that

\begin{eqnarray*}
&&E[y^{(t+1)}f(x^{(t+1)};w_i^{(t)})|\mathcal{F}_t]\\
&=&\frac{1}{2}E[max\{0,\langle x_{d+1\dots 2d}^{(t+1)},w_{\perp,i}^{(t)}\rangle,\dots,\langle x_{D-d+1\dots D}^{(t+1)},w_{\perp,i}^{(t)}\rangle\}|\mathcal{F}_t,y^{(t+1)}=1]\\
&-&\frac{1}{2}E[max\{-{\alpha}_i^{(t)},\langle x_{d+1\dots 2d}^{(t+1)},w_{\perp,i}^{(t)}\rangle,\dots,\langle x_{D-d+1\dots D}^{(t+1)},w_{\perp,i}^{(t)}\rangle\}|\mathcal{F}_t,y^{(t+1)}=-1]\\
&=&\frac{1}{2}E[max\{0,\langle x_{d+1\dots 2d}^{(t+1)},w_{\perp,i}^{(t)}\rangle,\dots,\langle x_{D-d+1\dots D}^{(t+1)},w_{\perp,i}^{(t)}\rangle\}|\mathcal{F}_t]\\
&-&\frac{1}{2}E[max\{-{\alpha}_i^{(t)},\langle x_{d+1\dots 2d}^{(t+1)},w_{\perp,i}^{(t)}\rangle,\dots,\langle x_{D-d+1\dots D}^{(t+1)},w_{\perp,i}^{(t)}\rangle\}|\mathcal{F}_t]\\
&\le&0
\end{eqnarray*}

So according to the dynamics of $w_i^{(t)}$, we have that

\begin{eqnarray*}
&&E[{\|w_i^{(t+1)}\|}_2^2|\mathcal{F}_t]\\
&=&{\|w_i^{(t)}\|}_2^2+{\eta}^2I_{\{f(x^{(t+1)};w_i^{(t)})>0\}}+2\eta E[y^{(t+1)}f(x^{(t+1)};w_i^{(t)})|\mathcal{F}_t]\\
&\le&{\|w_i^{(t)}\|}_2^2+{\eta}^2
\end{eqnarray*}

combine with ${\|w_i^{(t)}\|}_2^2={({\alpha}_i^{(t)})}^2+{\|w_{\perp,i}^{(t)}\|}_2^2$ we have that

\begin{eqnarray*}
E[{\|w_{\perp,i}^{(t+1)}\|}_2^2|\mathcal{F}_t]&\le&{\|w_{\perp,i}^{(t)}\|}_2^2+{\eta}^2+{({\alpha}_i^{(t)})}^2-{({\alpha}_i^{(t+1)})}^2\\
\end{eqnarray*}

combine with $\alpha_i^{(t)}$'s monotonically increasing property, we have that

if ${\alpha}_i^{(0)}>0$, then

\begin{eqnarray*}
E({\|w_{\perp,i}^{(t)}\|}_2^2)&\le&{\|w_{\perp,i}^{(0)}\|}_2^2+\sum_{s=1}^{t}{\eta}^2\\
&=&{\|w_{\perp,i}^{(0)}\|}_2^2+{\eta}^2 t
\end{eqnarray*}

if ${\alpha}_i^{(0)}<0$, then

\begin{eqnarray*}
E({\|w_{\perp,i}^{(t)}\|}_2^2)&\le&{\|w_{\perp,i}^{(0)}\|}_2^2+{({\alpha}_i^{(0)})}^2+\sum_{s=1}^{t}{\eta}^2\\
&=&{\|w_{\perp,i}^{(0)}\|}_2^2+{({\alpha}_i^{(0)})}^2+{\eta}^2 t
\end{eqnarray*}

So no matter what ${\alpha}_i^{(0)}$ is, we always have that

\begin{eqnarray*}
E({\|w_{\perp,i}^{(t)}\|}_2^2)\le{\|w_{\perp,i}^{(0)}\|}_2^2+{({\alpha}_i^{(0)})}^2+{\eta}^2 t
\end{eqnarray*}

So we have that

\begin{eqnarray*}
E({\|w_{\perp,i}^{(t)}\|}_2)&\le&\sqrt{E({\|w_{\perp,i}^{(t)}\|}_2^2)}\\
&=&\sqrt{{\|w_{\perp,i}^{(0)}\|}_2^2+{({\alpha}_i^{(0)})}^2+{\eta}^2 t}\\
&\le&{\|w_{\perp,i}^{(0)}\|}_2+|{\alpha}_i^{(0)}|+\eta\sqrt{t}
\end{eqnarray*}

According to Markov's Inequality, we have that for any $\sigma>0$,

\begin{eqnarray*}
P\{{\|w_{\perp,i}^{(t)}\|}_2\ge t^{\frac{1}{2}+\sigma}\}&\le&\frac{{\|w_{\perp,i}^{(0)}\|}_2+|{\alpha}_i^{(0)}|+\eta\sqrt{t}}{t^{\frac{1}{2}+\sigma}}\\
&\le&(\eta+2)t^{-\sigma}
\end{eqnarray*}

(b.2)When $a_i=-1$, similar to case(b.1), we can prove that

\begin{eqnarray*}
P\{{\|w_{\perp,i}^{(t)}\|}_2\ge t^{\frac{1}{2}+\sigma}\}\le(\eta+2)t^{-\sigma}
\end{eqnarray*}

(c) Next we will prove the convergence rate of ${\theta}_t$.

(c.1)When $a_i=1$

Note that $\sin{\theta}_t=\frac{{\|w_{\perp,i}^{(t)}\|}_2}{{\|w_i^{(t)}\|}_2}\le \frac{{\|w_{\perp,i}^{(t)}\|}_2}{{\alpha}_i^{(t)}}$, so we have that

\begin{eqnarray*}
P\{\sin{\theta}_t\le\frac{t^{\frac{1}{2}+\sigma}}{{\alpha}_i^{(0)}+2^{-k-1}\eta t}\}&\ge&P\{\frac{{\|w_{\perp,i}^{(t)}\|}_2}{{\alpha}_i^{(t)}}\le\frac{t^{\frac{1}{2}+\sigma}}{{\alpha}_i^{(0)}+2^{-k-1}\eta t}\}\\
&\ge&P\{{\|w_{\perp,i}^{(t)}\|}_2\le t^{\frac{1}{2}+\sigma},{\alpha}_i^{(t)}\ge{\alpha}_i^{(0)}+2^{-k-1}\eta t\}\\
&\ge&P\{{\|w_{\perp,i}^{(t)}\|}_2\le t^{\frac{1}{2}+\sigma}\}+P\{{\alpha}_i^{(t)}\ge{\alpha}_i^{(0)}+2^{-k-1}\eta t\}-1\\
&\ge&1-(\eta+2)t^{-\sigma}-\frac{2^{k+2}-4}{t}
\end{eqnarray*}

(c.2)When $a_i=-1$

Similar to case(c.1), we can prove that

\begin{eqnarray*}
P\{\sin{\theta}_t\le\frac{t^{\frac{1}{2}+\sigma}}{-{\alpha}_i^{(0)}+2^{-k-1}\eta t}\}\ge1-(\eta+2)t^{-\sigma}-\frac{2^{k+2}-4}{t}
\end{eqnarray*}

(d) Finally, we will analyze the classification accuracy of the learned CNN over $\Dcal$.

With probability of $1-2\cdot{(\frac{1}{2})}^h$, there exist $i_1,i_2\in [h]$, s.t.

\begin{eqnarray*}
&&a_{i_1}=1\\
&&a_{i_2}=-1
\end{eqnarray*}

When $t\ge max\{\frac{(2^{k+2}-4)(h+2)}{\epsilon},{(\frac{(\eta+2)(h+2)}{\epsilon})}^{\frac{1}{\sigma}},{(\frac{2^{k+1}(2h+1)}{\eta})}^{\frac{2}{1-2\sigma}}\}$, we have that

\begin{eqnarray*}
P\{{\alpha}_{i_1}^{(t)}\le {\alpha}_{i_1}^{(0)}+2^{-k-1}\eta t\}&\le& \frac{\epsilon}{h+2}\\
P\{{\alpha}_{i_2}^{(t)}\ge {\alpha}_{i_2}^{(0)}-2^{-k-1}\eta t\}&\le& \frac{\epsilon}{h+2}\\
P\{{\|w_{\perp,i}^{(t)}\|}_2\ge t^{\frac{1}{2}+\sigma}\}&\le& \frac{\epsilon}{h+2},\ \forall i\in [h]
\end{eqnarray*}

So we have that with probability at least $1-2\cdot{(\frac{1}{2})}^h-\epsilon$

\begin{eqnarray*}
{\alpha}_{i_1}^{(t)}&\ge& {\alpha}_{i_1}^{(0)}+2^{-k-1}\eta t,\ a_{i_1}=1\\
{\alpha}_{i_2}^{(t)}&\le& {\alpha}_{i_2}^{(0)}-2^{-k-1}\eta t,\ a_{i_2}=-1\\
{\|w_{\perp,i}^{(t)}\|}_2&\le& t^{\frac{1}{2}+\sigma},\ \forall i\in [h]
\end{eqnarray*}

So we have that for almost all $(x,y)$ sampled from the clean distribution $\mathcal{D}$ (without loss of generality, we assume that $y=1$)

\begin{eqnarray*}
yF(x;W^{(t)})&=&\sum_{i=1}^h a_i f(x;w_i^{(t)})\\
&=&\sum_{i:a_i=1}f(x;w_i^{(t)})-\sum_{i:a_i=-1}f(x;w_i^{(t)})\\
&\ge&f(x;w_{i_1}^{(t)})-\sum_{i:a_i=-1}f(x;w_i^{(t)})\\
&=&max\{0,{\alpha}_{i_1}^{(t)},f(x;w_{\perp,i_1}^{(t)})\}-\sum_{i:a_i=-1}max\{0,{\alpha}_i^{(t)},f(x;w_{\perp,i}^{(t)})\}\\
&\ge&{\alpha}_{i_1}^{(t)}-\sum_{i:a_i=-1}(max\{0,{\alpha}_i^{(0)}\}+{\|w_{\perp,i}^{(t)}\|}_2)\\
&\ge&2^{-k-1}\eta t-h(1+t^{\frac{1}{2}+\sigma})\\
&\ge&t^{\frac{1}{2}+\sigma}\\
&\ge&1
\end{eqnarray*}

The second inequality is because $f(x;w_{\perp,i}^{(t)})\le {\|w_{\perp,i}^{(t)}\|}_2$ and when $a_i=-1$, ${\alpha}_i^{(t)}$ is monotonically decreasing.

The third inequality is because that $|{\alpha}_i^{(0)}|\le1$.

(2)When $\langle p_+,p_-\rangle\le0$ and $p_+\ne -p_-$,

Consider the decomposition of $w_i^{(t)}$

\begin{eqnarray*}
w_i^{(t)}=w_{\parallel,i}^{(t)}+w_{\perp,i}^{(t)}
\end{eqnarray*}

where $w_{\parallel,i}^{(t)}\in span\{p_+,p_-\}$ and $w_{\perp,i}^{(t)}$ is perpendicular to $p_+$ and $p_-$.

(a) First we will analyze the property of $w_{\parallel,i}^{(t)}$.

Without loss of generality, we assumed that $a_i=1$.

As Figure 6(a) shows, $\overrightarrow{OA}$ corresponds to $p_+$, $\overrightarrow{OB}$ corresponds to $p_-$, $\overrightarrow{OW}$ corresponds to $w_{\parallel,i}^{(t)}$, $\phi$ corresponds to ${\phi}_t$(the angle between $p_+$ and $w_{\parallel,i}^{(t)}$), line segment CD is perpendicular to $\overrightarrow{OA}$, line segment EF is perpendicular to $\overrightarrow{OB}$.

(a.1)When ${\phi}_0\in[0,\angle AOF]$,

If ${\phi}_t\in[0,\angle AOF]$, then $\langle p_+,w_{\parallel,i}^{(t)}\rangle>0$, $\langle p_-,w_{\parallel,i}^{(t)}\rangle\le0$, and the dynamics of $w_{\parallel,i}^{(t)}$ will be

\begin{eqnarray*}
w_{\parallel,i}^{(t+1)}=w_{\parallel,i}^{(t)}+\eta p_+\Ibb_{\{y^{(t+1)}=1,\langle p_+,w_{\parallel,i}^{(t)}\rangle>f(x^{(t+1)};w_{\perp,i}^{(t)})\}}
\end{eqnarray*}

Note that ${\phi}_t$ will always decrease when $w_{\parallel,i}^{(t)}$ is updated by $p_+$, so ${\phi}_t$ will monotonically decrease and will always be no less than $0$.

Let $z_t$ be the number of times that event "$y^{(s)}=1,f(x^{(s)};w_{\perp,i}^{(s-1)})=0$" holds, $s\in[t]$, then $z_t$ will be a lower bound of the number of times that $w_{\parallel,i}$ be updated by $p_+$ in the fist $t$ steps, so we have that

\begin{eqnarray*}
\sin{\phi}_t&\le&\frac{{\|w_{\parallel,i}^{(0)}\|}_2\sin{\phi}_0}{\sqrt{{\|w_{\parallel,i}^{(0)}\|}_2^2{\sin}^2{\phi}_0+{(\eta z_t)}^2}}\\
&\le&\frac{{\|w_{\parallel,i}^{(0)}\|}_2}{\sqrt{{\|w_{\parallel,i}^{(0)}\|}_2^2+{(\eta z_t)}^2}}
\end{eqnarray*}

Note that

\begin{eqnarray*}
E[\Ibb_{\{y^{(s)}=1,f(x^{(s)};w_{\perp,i}^{(s-1)})=0\}}|\mathcal{F}_{s-1}]=2^{-k}
\end{eqnarray*}

So $z_t\sim Binomial(t,2^{-k})$. According to Chebyshev's inequality, we have that

\begin{eqnarray*}
P\{z_t\le 2^{-k-1}t\}&\le&P\{|z_t-2^{-k}t|\ge 2^{-k-1}t\}\\
&\le&\frac{2^{-k}(1-2^{-k})t}{{(2^{-k-1}t)}^2}\\
&=&\frac{2^{k+2}-4}{t}
\end{eqnarray*}

So we have that

\begin{eqnarray*}
P\{\sin{\phi}_t\le\frac{{\|w_{\parallel,i}^{(0)}\|}_2}{\sqrt{{\|w_{\parallel,i}^{(0)}\|}_2^2+{(2^{-k-1}\eta t)}^2}}\}&\ge&P\{z_t\ge 2^{-k-1}t\}\\
&\ge&1-\frac{2^{k+2}-4}{t}
\end{eqnarray*}

According to the dynamics of $w_{\parallel,i}^{(t)}$, we also have that

\begin{eqnarray*}
\langle p_+,w_{\parallel,i}^{(t+1)}\rangle&=&\langle p_+,w_{\parallel,i}^{(t)}\rangle+\eta \Ibb_{\{y^{(t+1)}=1,\langle p_+,w_{\parallel,i}^{(t)}\rangle>f(x^{(t+1)};w_{\perp,i}^{(t)})\}}\\
&\ge&\langle p_+,w_{\parallel,i}^{(t)}\rangle+\eta \Ibb_{\{y^{(t+1)}=1,f(x^{(t+1)};w_{\perp,i}^{(t)})=0\}}
\end{eqnarray*}

So we have that

\begin{eqnarray*}
\langle p_+,w_{\parallel,i}^{(t)}\rangle&\ge&\langle p_+,w_{\parallel,i}^{(0)}\rangle+\eta z_t\\
&\ge&\eta z_t
\end{eqnarray*}

Similarly, according to Chebyshev's inequality, we can prove that

\begin{eqnarray*}
P\{\langle p_+,w_{\parallel,i}^{(t)}\rangle\le2^{-k-1}\eta t\}&\le&P\{z_t\le2^{-k-1}t\}\\
&\le&\frac{2^{k+2}-4}{t}
\end{eqnarray*}

(a.2)When ${\phi}_0\in(\angle AOF,\angle AOB)$,

If ${\phi}_t\in(\angle AOF,\angle AOB)$, then $\langle p_-,w_{\parallel,i}^{(t)}\rangle>0$ and the dynamics of $w_{\parallel,i}^{(t)}$ will be

\begin{eqnarray*}
w_{\parallel,i}^{(t+1)}&=&w_{\parallel,i}^{(t)}+\eta p_+\Ibb_{\{y^{(t+1)}=1,\langle p_+,w_{\parallel,i}^{(t)}\rangle>f(x^{(t+1)};w_{\perp,i}^{(t)})\}}\\
&-&\eta p_-\Ibb_{\{y^{(t+1)}=-1,\langle p_-,w_{\parallel,i}^{(t)}\rangle>f(x^{(t+1)};w_{\perp,i}^{(t)})\}}
\end{eqnarray*}

Note that ${\phi}_t$ will always decrease when $w_{\parallel,i}^{(t)}$ is updated by $p_+$ or $p_-$, so ${\phi}_t$ will monotonically decrease.

Before ${\phi}_t$ decreases below to $\angle AOF$, we always have that

\begin{eqnarray*}
\langle p_-,w_{\parallel,i}^{(t+1)}\rangle&=&\langle p_-,w_{\parallel,i}^{(t)}\rangle+\eta\langle p_+,p_-\rangle \Ibb_{\{y^{(t+1)}=1,\langle p_+,w_{\parallel,i}^{(t)}\rangle>f(x^{(t+1)};w_{\perp,i}^{(t)})\}}\\
&-&\eta \Ibb_{\{y^{(t+1)}=-1,\langle p_-,w_{\parallel,i}^{(t)}\rangle>f(x^{(t+1)};w_{\perp,i}^{(t)})\}}\\
&\le&\langle p_-,w_{\parallel,i}^{(t)}\rangle
\end{eqnarray*}

So before ${\phi}_t$ decreases below to $\angle AOF$, we have that $\langle p_-,w_{\parallel,i}^{(t)}\rangle$ is monotonically decreasing.

(a.3)When ${\phi}_0=\angle AOB$

If ${\phi}_t=\angle AOB$, then the dynamics of $w_{\parallel,i}^{(t)}$ will be

\begin{eqnarray*}
w_{\parallel,i}^{(t+1)}=w_{\parallel,i}^{(t)}-\eta p_-\Ibb_{\{y^{(t+1)}=-1,\langle p_-,w_{\parallel,i}^{(t)}\rangle>f(x^{(t+1)};w_{\perp,i}^{(t)})\}}
\end{eqnarray*}

So the norm of $w_{\parallel,i}^{(t)}$ will monotonically decrease and ${\phi}_t=\angle AOB$ will always hold before $w_{\parallel,i}^{(t)}$ changes its direction. So before $w_{\parallel,i}^{(t)}$ changes its direction, we have that $\langle p_-,w_{\parallel,i}^{(t)}\rangle$ is monotonically decreasing.

(a.4)When ${\phi}_0\in(\angle AOB,\angle AOE)$

If ${\phi}_t\in(\angle AOB,\angle AOE)$, then the dynamics of $w_{\parallel,i}^{(t)}$ will be

\begin{eqnarray*}
w_{\parallel,i}^{(t+1)}=w_{\parallel,i}^{(t)}-\eta p_-\Ibb_{\{y^{(t+1)}=-1,\langle p_-,w_{\parallel,i}^{(t)}\rangle>f(x^{(t+1)};w_{\perp,i}^{(t)})\}}
\end{eqnarray*}

Then ${\phi}_t$ will monotonically increase. And similar to case(a.2) before ${\phi}_t$ increases above to $\angle AOE$, we have that $\langle p_-,w_{\parallel,i}^{(t)}\rangle$ is monotonically decreasing.

(a.5)When ${\phi}_0\in[\angle AOE,\frac{3\pi}{2}]$

If ${\phi}_t\in[\angle AOE,\frac{3\pi}{2}]$, then $\langle p_+,w_{\parallel,i}^{(t)}\rangle\le0$ and $\langle p_-,w_{\parallel,i}^{(t)}\rangle\le0$, so $w_{\parallel,i}^{(t)}$ will not be updated.

(a.6)When ${\phi}_0\in(\frac{3\pi}{2},2\pi]$

Similar to case(a.1), we can proof that

\begin{eqnarray*}
P\{\sin(2\pi-{\phi}_t)\le\frac{{\|w_{\parallel,i}^{(0)}\|}_2}{\sqrt{{\|w_{\parallel,i}^{(0)}\|}_2^2+{(2^{-k-1}\eta t)}^2}}\}\ge1-\frac{2^{k+2}-4}{t}
\end{eqnarray*}

and

\begin{eqnarray*}
P\{\langle p_+,w_{\parallel,i}^{(t)}\rangle\le2^{-k-1}\eta t\}\le\frac{2^{k+2}-4}{t}
\end{eqnarray*}

(b) Next we will analyze the property of $w_{\perp,i}^{(t)}$

Without loss of generality, we assumed that $a_i=1$.

When $\langle p_+,w_{\parallel,i}^{(t)}\rangle\ge\langle p_-,w_{\parallel,i}^{(t)}\rangle$, we have that $max\{0,\langle p_+,w_{\parallel,i}^{(t)}\rangle\}\ge max\{0,\langle p_-,w_{\parallel,i}^{(t)}\rangle\}$ and the dynamics of $w_{\perp,i}^{(t)}$ is

\begin{eqnarray*}
w_{\perp,i}^{(t+1)}&=&w_{\perp,i}^{(t)}+\eta x_{u_1d+1\dots(u_1+1)d}^{(t+1)}\Ibb_{\{y^{(t+1)}=1,max\{0,\langle p_+,w_{\parallel,i}^{(t)}\rangle\}<f(x^{(t+1)};w_{\perp,i}^{(t)})\}}\\
&-&\eta x_{u_2d+1\dots(u_2+1)d}^{(t+1)}\Ibb_{\{y^{(t+1)}=-1,max\{0,\langle p_-,w_{\parallel,i}^{(t)}\rangle\}<f(x^{(t+1)};w_{\perp,i}^{(t)})\}}\\
\end{eqnarray*}

where $u_1,u_2\in[k]$ and $u_1,u_2\ne1$.

Similar to theorem1(1)(b.1), we can proof that

\begin{eqnarray*}
E[{\|w_{\perp,i}^{(t+1)}\|}_2^2|\mathcal{F}_t]\le {\|w_{\perp,i}^{(t)}\|}_2^2+{\eta}^2
\end{eqnarray*}

When $\langle p_+,w_{\parallel,i}^{(t)}\rangle\le\langle p_-,w_{\parallel,i}^{(t)}\rangle$, we have that $max\{0,\langle p_+,w_{\parallel,i}^{(t)}\rangle\}\le max\{0,\langle p_-,w_{\parallel,i}^{(t)}\rangle\}$.

According to the dynamics of $w_i^{(t)}$, we have that

\begin{eqnarray*}
E[{\|w_i^{(t+1)}\|}_2^2|\mathcal{F}_t]&\le&{\|w_i^{(t)}\|}_2^2+{\eta}^2+2\eta E[y^{(t+1)}f(x^{(t+1)};w_i^{(t)})|\mathcal{F}_t]\\
&=&{\|w_i^{(t)}\|}_2^2+{\eta}^2+2\eta E[max\{max\{0,\langle p_+,w_{\parallel,i}^{(t)}\rangle\},f(x^{(t+1)};w_{\perp,i}^{(t)})\}\Ibb_{\{y^{(t+1)}=1\}}|\mathcal{F}_t]\\
&-&2\eta E[max\{max\{0,\langle p_-,w_{\parallel,i}^{(t)}\rangle\},f(x^{(t+1)};w_{\perp,i}^{(t)})\}\Ibb_{\{y^{(t+1)}=-1\}}|\mathcal{F}_t]\\
&\le&{\|w_i^{(t)}\|}_2^2+{\eta}^2
\end{eqnarray*}

Similar to theorem1(1)(b.1), we can proof that

\begin{eqnarray*}
E[{\|w_{\perp,i}^{(t+1)}\|}_2^2|\mathcal{F}_t]\le {\|w_{\perp,i}^{(t)}\|}_2^2+{\eta}^2+{\|w_{\parallel,i}^{(t)}\|}_2^2-{\|w_{\parallel,i}^{(t+1)}\|}_2^2
\end{eqnarray*}

According to the proof of theorem1(2)(a), we know that $\langle p_+-p_-,w_{\parallel,i}^{(t)}\rangle$ is monotonous. So similar to theorem1(1)(b.1), we can proof that for any $\sigma>0$

\begin{eqnarray*}
P\{{\|w_{\perp,i}^{(t)}\|}_2\ge t^{\frac{1}{2}+\sigma}\}\le(\eta+2)t^{-\sigma}
\end{eqnarray*}

(c) Next we will prove the convergence rate of ${\theta}_t$.

Without loss of generality, we assumed that $a_i=1$.

When ${\phi}_0\in[0,\angle AOF]$, according theorem1(2)(a.1) and theorem1(2)(b), we have that

\begin{eqnarray*}
P\{\sin{\phi}_t\le\frac{{\|w_{\parallel,i}^{(0)}\|}_2}{\sqrt{{\|w_{\parallel,i}^{(0)}\|}_2^2+{(2^{-k-1}\eta t)}^2}}\}&\ge&1-\frac{2^{k+2}-4}{t}\\
P\{\langle p_+,w_{\parallel,i}^{(t)}\rangle\le2^{-k-1}\eta t\}&\le&\frac{2^{k+2}-4}{t}\\
P\{{\|w_{\perp,i}^{(t)}\|}_2\ge t^{\frac{1}{2}+\sigma}\}&\le&(\eta+2)t^{-\sigma}\\
\end{eqnarray*}

Note that

\begin{eqnarray*}
\sin{\theta}_t&\le&\frac{\sqrt{{\|w_{\parallel,i}^{(t)}\|}_2^2{sin}^2{\phi}_t+{\|w_{\perp,i}^{(t)}\|}_2^2}}{{\|w_i^{(t)}\|}_2}\\
&\le&\frac{{\|w_{\parallel,i}^{(t)}\|}_2sin{\phi}_t+{\|w_{\perp,i}^{(t)}\|}_2}{\langle p_+,w_{\parallel,i}^{(t)}\rangle}
\end{eqnarray*}

and

\begin{eqnarray*}
{\|w_{\parallel,i}^{(t)}\|}_2\le{\|w_{\parallel,i}^{(0)}\|}_2+\eta t
\end{eqnarray*}

So similar to theorem1(1)(c.1), we can prove that for any $\sigma>0$, we have that

\begin{eqnarray*}
P\{\sin{\theta}_t\le\frac{t^{\frac{1}{2}+\sigma}+\frac{({\|w_{\parallel,i}^{(0)}\|}_2+\eta t){\|w_{\parallel,i}^{(0)}\|}_2}{\sqrt{{\|w_{\parallel,i}^{(0)}\|}_2^2+{(2^{-k-1}\eta t)}^2}}}{2^{-k-1}\eta t}\}\ge1-\frac{2^{k+3}-8}{t}-(\eta+2)t^{-\sigma}
\end{eqnarray*}

Similarly, when ${\phi}_0\in(\frac{3\pi}{2},2\pi]$, we also have that

\begin{eqnarray*}
P\{sin{\theta}_t\le\frac{t^{\frac{1}{2}+\sigma}+\frac{({\|w_{\parallel,i}^{(0)}\|}_2+\eta t){\|w_{\parallel,i}^{(0)}\|}_2}{\sqrt{{\|w_{\parallel,i}^{(0)}\|}_2^2+{(2^{-k-1}\eta t)}^2}}}{2^{-k-1}\eta t}\}\ge1-\frac{2^{k+3}-8}{t}-(\eta+2)t^{-\sigma}
\end{eqnarray*}

(d) Finally, we will analyze the classification accuracy of the learned CNN over $\Dcal$.

Let $\rho=\frac{\angle AOF+\frac{\pi}{2}}{2\pi}\ge\frac{1}{4}$, then with probability of $1-2\cdot{(1-\frac{\rho}{2})}^h\ge1-2\cdot{(\frac{7}{8})}^h$, there exist $i_1,i_2\in[h]$, s.t.

\begin{eqnarray*}
&&a_{i_1}=1,the\ angle\ between\ p_+\ and\ w_{\parallel,i_1}^{(0)}\in (-\frac{\pi}{2},\angle AOF]\\
&&a_{i_2}=-1,the\ angle\ between\ p_+\ and\ w_{\parallel,i_2}^{(0)}\in [\frac{\pi}{2},\frac{\pi}{2}+\angle AOB)\\
\end{eqnarray*}

When $t\ge max\{\frac{(2^{k+2}-4)(h+2)}{\epsilon},{(\frac{(\eta+2)(h+2)}{\epsilon})}^{\frac{1}{\sigma}},{(\frac{2^{k+1}(2h+1)}{\eta})}^{\frac{2}{1-2\sigma}}\}$, we have that

\begin{eqnarray*}
P\{\langle p_+,w_{\parallel,i_1}^{(t)}\rangle\le2^{-k-1}\eta t\}&\le&\frac{\epsilon}{h+2}\\
P\{\langle p_-,w_{\parallel,i_2}^{(t)}\rangle\le2^{-k-1}\eta t\}&\le&\frac{\epsilon}{h+2}\\
P\{{\|w_{\perp,i}^{(t)}\|}_2\ge t^{\frac{1}{2}+\sigma}\}&\le&\frac{\epsilon}{h+2},\ i\in[h]
\end{eqnarray*}

So we have that with probability at least $1-2\cdot{(\frac{7}{8})}^h-\epsilon$

\begin{eqnarray*}
\langle p_+,w_{\parallel,i_1}^{(t)}\rangle&\le&2^{-k-1}\eta t\\
\langle p_-,w_{\parallel,i_2}^{(t)}\rangle&\le&2^{-k-1}\eta t\\
{\|w_{\perp,i}^{(t)}\|}_2&\ge&t^{\frac{1}{2}+\sigma},\ i\in[h]
\end{eqnarray*}

So similar to theorem1(1)(d) we have that for almost all $(x,y)$ sampled from the clean distribution $\Dcal$ (without loss of generality, we assume that $y=1$)

\begin{eqnarray*}
yF(x;W^{(t)})&=&\sum_{i=1}^ha_if(x;w_i^{(t)})\\
&\ge&\langle p_+,w_{\parallel,i_1}^{(t)}\rangle-\sum_{i:a_i=-1}(max\{0,\langle p_+,w_{\parallel,i}^{(0)}\rangle\}+{\|w_{\perp,i}^{(t)}\|}_2)\\
&\ge&1
\end{eqnarray*}

The first inequality is because that when $a_i=-1$, $\langle p_+,w_{\parallel,i}^{(t)}\rangle$ is monotonically decreasing.

(3)When $\langle p_+,p_-\rangle>0$, $p_+\ne p_-$, the positive examples and negative examples sample alternatively.

Consider the decomposition of $w_i^{(t)}$

\begin{eqnarray*}
w_i^{(t)}=w_{\parallel,i}^{(t)}+w_{\perp,i}^{(t)}
\end{eqnarray*}

where $w_{\parallel,i}^{(t)}\in$ span$\{p_+,p_-\}$ and $w_{\perp,i}^{(t)}$ is perpendicular to $p_+$ and $p_-$.

(a) First we will analyze the property of $w_{\parallel,i}^{(t)}$ and $w_{\perp,i}^{(t)}$.

Without loss of generality, we assume that $a_i=1$.

As Figure 6(b) shows, $\overrightarrow{OA}$ corresponds to $p_+$, $\overrightarrow{OB}$ corresponds to $p_-$, $\overrightarrow{OW}$ corresponds to $w_{\parallel,i}^{(t)}$, $\phi$ corresponds to ${\phi}_t$(the angle between $p_-$ and $w_{\parallel,i}^{(t)}$), line segment CD is perpendicular to $\overrightarrow{OA}$, line segment EF is perpendicular to $\overrightarrow{OB}$. And let $\hat{\phi}=\angle BOA$.

Let $T$ be a positive even integer.

(a.1)When ${\phi}_T\in[\hat{\phi},\angle BOC)$

Similar to theorem1(1) and theorem1(2), we can prove that for any $\sigma>0$

\begin{eqnarray*}
P\{{\|w_{\parallel,i}^{(T)}\|}_2-{\|w_{\perp,i}^{(T)}\|}_2\ge\Omega(T)\}\ge1-O(T^{-\sigma})
\end{eqnarray*}

If in the $(T+2t+1)$-th step, $p_+$ is activated by $w_i$ and in the $(T+2t+2)$-th step, a non-key pattern is activated by $w_i$, then we have that

\begin{eqnarray*}
w_{\parallel,i}^{(T+2t+2)}&=&w_{\parallel,i}^{(T+2t+1)}=w_{\parallel,i}^{(T+2t)}+\eta p_+\\
w_{\perp,i}^{(T+2t+2)}&=&w_{\perp,i}^{(T+2t+1)}-\eta x_{ud+1...(u+1)d}^{(T+2t+2)}=w_{\perp,i}^{(T+2t)}-\eta x_{ud+1...(u+1)d}^{(T+2t+2)}
\end{eqnarray*}

where $\langle x_{ud+1...(u+1)d}^{(T+2t+2)},w_{\perp,i}^{(T+2t+1)}\rangle>0$.

Then we have that

\begin{eqnarray*}
{\|w_{\parallel,i}^{(T+2t+2)}\|}_2&=&{\|w_{\parallel,i}^{(T+2t+1)}\|}_2\ge{\|w_{\parallel,i}^{(T+2t)}\|}_2+\eta sin\hat{\phi}\\
{\|w_{\perp,i}^{(T+2t+2)}\|}_2^2&\le&{\|w_{\perp,i}^{(T+2t+1)}\|}_2^2+{\eta}^2={\|w_{\perp,i}^{(T+2t)}\|}_2^2+{\eta}^2
\end{eqnarray*}

If in the $(T+2t+1)$-th step, $p_+$ is activated by $w_i$ and in the $(T+2t+2)$-th step, $p_-$ is activated by $w_i$, then we have that

\begin{eqnarray*}
w_{\parallel,i}^{(T+2t+2)}&=&w_{\parallel,i}^{(T+2t+1)}-\eta p_-=w_{\parallel,i}^{(T+2t)}+\eta(p_+-p_-)\\
w_{\perp,i}^{(T+2t+2)}&=&w_{\perp,i}^{(T+2t+1)}=w_{\perp,i}^{(T+2t)}
\end{eqnarray*}

Then we have that

\begin{eqnarray*}
{\|w_{\parallel,i}^{(T+2t+2)}\|}_2\ge{\|w_{\parallel,i}^{(T+2t)}\|}_2+\eta sin\frac{\hat{\phi}}{2}\\
{\|w_{\perp,i}^{(T+2t+2)}\|}_2={\|w_{\perp,i}^{(T+2t)}\|}_2
\end{eqnarray*}

When $T$ is sufficiently large, we can see that ${\|w_{\parallel,i}^{(T+t)}\|}_2sin\hat{\phi}\ge{\|w_{\perp,i}^{(T+t)}\|}_2$ always holds, so when positive example is encountered, $p_+$ must be activated, so we have that

\begin{eqnarray*}
{\|w_{\parallel,i}^{(T+t+1)}\|}_2&\ge& {\|w_{\parallel,i}^{(T+t)}\|}_2+\frac{\eta}{2}sin\frac{\hat{\phi}}{2}\\
{\|w_{\perp,i}^{(T+t+1)}\|}_2^2&\le&{\|w_{\perp,i}^{(T+t)}\|}_2^2+{\eta}^2
\end{eqnarray*}

always hold. So that

\begin{eqnarray*}
{\|w_{\parallel,i}^{(T+t)}\|}_2&\ge&{\|w_{\parallel,i}^{(T)}\|}_2+\frac{\eta}{2}sin\frac{\hat{\phi}}{2}t\\
{\|w_{\perp,i}^{(T+t)}\|}_2&\le&{\|w_{\perp,i}^{(T)}\|}_2+\eta\sqrt{t}
\end{eqnarray*}

Note that if $\langle p_-,w_{\parallel,i}^{(T+2t)}\rangle>{\|w_{\perp,i}^{(T+2t)}\|}_2$, then at $(T+2t+2)$-th step $p_-$ must be activated, and then

\begin{eqnarray*}
\langle p_-,w_{\parallel,i}^{(T+2t+2)}\rangle&=&\langle p_-,w_{\parallel,i}^{(T+2t)}\rangle+\eta(\langle p_+,p_-\rangle-1)\\
&=&\langle p_-,w_{\parallel,i}^{(T+2t)}\rangle-\eta(1-cos\hat{\phi})\\
{\|w_{\perp,i}^{(T+2t+2)}\|}_2&=&{\|w_{\perp,i}^{(T+2t)}\|}_2
\end{eqnarray*}

So we have that

\begin{eqnarray*}
\langle p_-,w_{\parallel,i}^{(T+2t+2)}\rangle-{\|w_{\perp,i}^{(T+2t+2)}\|}_2&\le&\langle p_-,w_{\parallel,i}^{(T+2t)}\rangle-{\|w_{\perp,i}^{(T+2t)}\|}_2-\eta(1-\cos\hat{\phi})
\end{eqnarray*}

So we have that

\begin{eqnarray*}
\langle p_-,w_{\parallel,i}^{(T+2t)}\rangle-{\|w_{\perp,i}^{(T+2t)}\|}_2&\le& max\{\langle p_-,w_{\parallel,i}^{(T)}\rangle-{\|w_{\perp,i}^{(T)}\|}_2,\eta\langle p_+,p_-\rangle\}\\
\end{eqnarray*}

Let $M=max\{\langle p_-,w_{\parallel,i}^{(T)}\rangle-{\|w_{\perp,i}^{(T)}\|}_2,\eta\langle p_+,p_-\rangle\}+\eta\langle p_+,p_-\rangle$, note that

\begin{eqnarray*}
\langle p_-,w_{\parallel,i}^{(T+2t+1)}\rangle-{\|w_{\perp,i}^{(T+2t+1)}\|}_2&\le&\langle p_-,w_{\parallel,i}^{(T+2t)}\rangle-{\|w_{\perp,i}^{(T+2t)}\|}_2+\eta\langle p_+,p_-\rangle\\
&\le&M
\end{eqnarray*}

So we have that

\begin{eqnarray*}
\langle p_-,w_{\parallel,i}^{(T+t)}\rangle-{\|w_{\perp,i}^{(T+t)}\|}_2\le M
\end{eqnarray*}

always holds for $t$.

So that

\begin{eqnarray*}
\frac{\langle p_-,w_{\parallel,i}^{(T+t)}\rangle}{{\|w_{\parallel,i}^{(T+t)}\|}_2}&\le&\frac{{\|w_{\parallel,i}^{(T+t)}\|}_2+M}{{\|w_{\parallel,i}^{(T+t)}\|}_2}\\
&\le&\frac{\eta\sqrt{t}+M}{\frac{\eta}{2}sin\frac{\hat{\phi}}{2}t}\\
&\to&0(t\to\infty)
\end{eqnarray*}

(a.2)When ${\phi}_T\in[\angle BOC,\frac{3\pi}{2}]$

Then $p_+$ and $p_-$ will not be activated, then $w_{\parallel,i}^{(T+t)}=w_{\parallel,i}^{(T)}$ will always hold.

(a.3)When ${\phi}_T\in(\frac{3\pi}{2},\frac{3\pi}{2}+\hat{\phi}]$

When ${\phi}_0\in(\frac{3\pi}{2},\frac{3\pi}{2}+\hat{\phi}]$, then ${\phi}_t$ will monotonically decreasing and similar to theorem1(1) and theorem1(2), we can prove that

\begin{eqnarray*}
P\{{\phi}_T\in(\frac{3\pi}{2},\frac{3\pi}{2}+\hat{\phi}]\}\le O(\frac{1}{T})
\end{eqnarray*}

(a.4)When ${\phi}_T\in(\frac{3\pi}{2}+\hat{\phi},2\pi]$

When ${\phi}_0\in(\frac{3\pi}{2}+\hat{\phi},2\pi]$, then ${\phi}_t$ will move out of this interval. Similar to theorem1(1) and theorem1(2), we can prove that

\begin{eqnarray*}
P\{{\phi}_T\in(\frac{3\pi}{2}+\hat{\phi},2\pi+\hat{\phi})\}\le O(\frac{1}{T})
\end{eqnarray*}

(a.5)When ${\phi}_T\in(0,\hat{\phi})$

When ${\phi}_0\in(0,\hat{\phi})$, then ${\phi}_t$ will monotonically increasing and similar to theorem1(1) and theorem1(2), we can prove that

\begin{eqnarray*}
P\{{\phi}_T\in(\frac{3\pi}{2}+\hat{\phi},2\pi+\hat{\phi})\}\le O(\frac{1}{T})
\end{eqnarray*}

(b) Next we will prove the convergence rate of $w_i^{(t)}$

Without loss of generality, we assume that $a_i=1$.

Let $p_*$ be a pattern that has the same direction as vector $\overrightarrow{OE}$ in Figure 1(b). Similar to Theorem1(1)(c) and Theorem1(2)(c), we can prove that for any $\sigma\in(0,\frac{1}{2})$

\begin{eqnarray*}
P\{\sin{\theta}_{T+t}\le O(\frac{1}{t^{\frac{1}{2}-\sigma}})\}\ge1-O(t^{-\sigma})-O(T^{-\sigma})
\end{eqnarray*}

(c) Finally, we will analyze the classification accuracy of the learned CNN over $\mathcal{D}$.

Similar to theorem1(1)(d) and theorem1(2)(d), we can prove that for any $\sigma>0$, $\epsilon>0$, when $T$ and $t$ are sufficiently large, with probability at least $1-2\cdot{(\frac{7}{8})}^h-\epsilon$

\begin{eqnarray*}
P_{(x,y)\sim\mathcal{D}}(yF(x;W^{(T+t)})\ge1)=1
\end{eqnarray*}

\paragraph{Theorem 2}
(\textbf{Population Loss with Noisy Data Distribution})
Assumed that $p_+=-p_-=p$,  $(x^{(t+1)},y^{(t+1)}),t=0,1,...$ are i.i.d. drawn from the $\epsilon$-noisy distribution $\mathcal{D}_{\epsilon}$, where $\epsilon$ satisfies $0<\epsilon<2^{-2k-1}$. Then

(1) For any $\sigma\in(0,\frac{1}{2})$, let ${\theta}_t$ be the angle between $a_ip$ and $w_i^{(t)}$, then
\begin{equation*}
  P\{\sin{\theta}_t\le O(t^{-(\frac{1}{2}-\sigma)})\}\ge1-O(t^{-\sigma})
\end{equation*}

(2) If $|\{i|a_i=1\}|=|\{i|a_i=-1\}|$, the classification accuracy over $\Dcal_{\epsilon}$ will be 1 when $t\to\infty$.

\paragraph{Proof}Without loss of generality, we assume that $u^*$ in definition 1 is equaled to $1$.

Consider the decomposition of $w_i^{(t)}$

\begin{eqnarray*}
w_i^{(t)}={\alpha}_i^{(t)}p+w_{\perp,i}^{(t)}
\end{eqnarray*}

where ${\alpha}_i^{(t)}\in \Rbb$ and $w_{\perp,i}^{(t)}$ is perpendicular to p.

Define some events

\begin{eqnarray*}
A_1&=&\{x_{1...d}^{(t+1)}\ is\ activated\ by\ w_i^{(t)}\}\\
A_2&=&\{x_{ud+1...(u+1)d}^{(t+1)}\ is\ activated\ by\ w_i^{(t)}\ for\ some\ u\ne1\}\\
\end{eqnarray*}

(a) First we will analyze the property of ${\alpha}_i^{(t)}$.

Without loss of generality, we assume that $a_i=1$.

According to the dynamics of $w_i^{(t)}$, we have that

\begin{eqnarray*}
{\alpha}_i^{(t+1)}p&=&{\alpha}_i^{(t)}p+\eta y^{(t+1)}x_{0,1...d}^{(t+1)}I_{A_1}+\eta y^{(t+1)}x_{1,ud+1...(u+1)d}^{(t+1)}I_{A_2}\\
&=&{\alpha}_i^{(t)}p+\eta pI_{A_1}+\eta y^{(t+1)}x_{1,ud+1...(u+1)d}^{(t+1)}I_{A_2}\\
\end{eqnarray*}

So we have that

\begin{eqnarray*}
{\alpha}_i^{(t+1)}&\ge&{\alpha}_i^{(t)}+\eta I_{A_1}-\eta\epsilon I_{A_2}\\
&\ge&{\alpha}_i^{(t)}+\eta I_{A_1}-\eta\epsilon
\end{eqnarray*}

Note that

\begin{eqnarray*}
\langle x_{1...d}^{(t+1)},w_i^{(t)}\rangle&=&\langle y^{(t+1)}p,{\alpha}_i^{(t)}p\rangle\\
\langle x_{ud+1...(u+1)d}^{(t+1)},w_i^{(t)}\rangle&=&\langle x_{1,ud+1...(u+1)d}^{(t+1)},{\alpha}_i^{(t)}p\rangle\\
&+&\langle x_{0,ud+1...(u+1)d}^{(t+1)}+x_{2,ud+1...(u+1)d}^{(t+1)},w_{\perp,i}^{(t)}\rangle,\ u\ne1
\end{eqnarray*}

So we have that

\begin{eqnarray*}
P(A_1|{\alpha}_i^{(t)}>0)&\ge&P(\{y^{(t+1)}=1\}\cap A_1|{\alpha}_i^{(t)}>0)\\
&\ge&P\{y^{(t+1)}=1,\langle x_{1,ud+1...(u+1)d}^{(t+1)},p\rangle<0,\\
&&\langle x_{0,ud+1...(u+1)d}^{(t+1)}+x_{2,ud+1...(u+1)d}^{(t+1)},w_{\perp,i}^{(t)}\rangle<0,\forall u\ne 1|{\alpha}_i^{(t)}>0\}\\
&=&P\{y^{(t+1)}=1,\langle x_{1,ud+1...(u+1)d}^{(t+1)},p\rangle<0,\\
&&\langle x_{0,ud+1...(u+1)d}^{(t+1)}+x_{2,ud+1...(u+1)d}^{(t+1)},w_{\perp,i}^{(t)}\rangle<0,\forall u\ne 1\}\\
&=&2^{-2k+1}
\end{eqnarray*}

Similarly we have that

\begin{eqnarray*}
P(A_1|{\alpha}_i^{(t)}<0)\ge 2^{-2k+1}
\end{eqnarray*}

that is to say no matter what ${\alpha}_i^{(t)}$ and $w_{\perp,i}^{(t)}$ are, ${\alpha}_i^{(t+1)}\ge {\alpha}_i^{(t)}+\eta-\eta\epsilon$ will be held with probability at least $2^{-2k+1}$. And ${\alpha}_i^{(t+1)}\ge{\alpha}_i^{(t)}-\eta\epsilon$ always holds.

So there exists a sequence of i.i.d. Bernoulli random variables $e_1,...,e_t$ satisfies $P\{e_s=1\}=2^{-2k+1},s=1,...,t$, and

\begin{eqnarray*}
{\alpha}_i^{(t)}\ge{\alpha}_i^{(0)}+\eta\sum_{s=1}^te_s-\eta\epsilon t
\end{eqnarray*}

So we have that

\begin{eqnarray*}
E({\alpha}_i^{(t)})&\ge&{\alpha}_i^{(0)}+2^{-2k+1}\eta t-\eta\epsilon t\\
&>&{\alpha}_i^{(0)}+2^{-2k}\eta t
\end{eqnarray*}

and

\begin{eqnarray*}
P\{{\alpha}_i^{(t)}\le{\alpha}_i^{(0)}+2^{-2k-1}\eta t\}&=&P\{{\alpha}_i^{(t)}-{\alpha}_i^{(0)}+\eta\epsilon t\le 2^{-2k-1}\eta t+\eta\epsilon t\}\\
&\le&P\{\sum_{s=1}^t e_s\le 2^{-2k-1}t+\epsilon t\}\\
&\le&P\{\sum_{s=1}^t e_s\le 2^{-2k}t\}\\
&\le&P\{|\sum_{s=1}^t e_s-2^{-2k+1}t|\ge 2^{-2k}t\}\\
&\le&\frac{2^{2k+1}-4}{t}
\end{eqnarray*}

(b) Next we will analyze the property of $w_{\perp,i}^{(t)}$.

Without loss of generality, we assume that $a_i=1$.

According to the dynamics of $w_i^{(t)}$, we have that

\begin{eqnarray*}
w_{\perp,i}^{(t+1)}=w_{\perp,i}^{(t)}+\eta y^{(t+1)}(x_{0,ud+1...(u+1)d}^{(t+1)}+x_{2,ud+1...(u+1)d}^{(t+1)})I_{A_2}
\end{eqnarray*}

If ${\alpha}_i^{(0)}>0$, we have that

\begin{eqnarray*}
E[{\|w_{\perp,i}^{(t+1)}\|}_2^2|\mathcal{F}_t]&=&{\|w_{\perp,i}^{(t)}\|}_2^2+{\eta}^2{\|x_{0,ud+1...(u+1)d}^{(t+1)}+x_{2,ud+1...(u+1)d}^{(t+1)}\|}_2^2E[I_{A_2}|\mathcal{F}_t]\\
&+&2\eta E[y^{(t+1)}\langle x_{0,ud+1...(u+1)d}^{(t+1)}+x_{2,ud+1...(u+1)d}^{(t+1)},w_{\perp,i}^{(t)}\rangle I_{A_2}|\mathcal{F}_t]\\
&\le&{\|w_{\perp,i}^{(t)}\|}_2^2+{\eta}^2{\|x_{0,ud+1...(u+1)d}^{(t+1)}+x_{2,ud+1...(u+1)d}^{(t+1)}\|}_2^2E[I_{A_2}|\mathcal{F}_t]\\
&\le&{\|w_{\perp,i}^{(t)}\|}_2^2+{\eta}^2{(1+\epsilon)}^2\\
&\le&{\|w_{\perp,i}^{(t)}\|}_2^2+{\eta}^2{(1+2\epsilon)}^2
\end{eqnarray*}

The first inequality is because that

\begin{eqnarray*}
P\{A_2|y^{(t+1)}=1\}\le P\{A_2|y^{(t+1)}=-1\}
\end{eqnarray*}

So that

\begin{eqnarray*}
&&E[y^{(t+1)}\langle x_{0,uk+1...(u+1)k}^{(t+1)}+x_{2,uk+1...(u+1)k}^{(t+1)},w_{\perp,i}^{(t)}\rangle I_{A_2}|\mathcal{F}_t]\\
&=&\frac{1}{2}E[\langle x_{0,uk+1...(u+1)k}^{(t+1)}+x_{2,uk+1...(u+1)k}^{(t+1)},w_{\perp,i}^{(t)}\rangle I_{A_2}|y^{(t+1)}=1,\mathcal{F}_t]\\
&-&\frac{1}{2}E[\langle x_{0,uk+1...(u+1)k}^{(t+1)}+x_{2,uk+1...(u+1)k}^{(t+1)},w_{\perp,i}^{(t)}\rangle I_{A_2}|y^{(t+1)}=-1,\mathcal{F}_t]\\
&\le&0
\end{eqnarray*}

If ${\alpha}_i^{(t)}<0$, we have that

\begin{eqnarray*}
E[{\|w_i^{(t+1)}\|}_2^2|\mathcal{F}_t]&\le&{\|w_i^{(t)}\|}_2^2+{\eta}^2{(1+2\epsilon)}^2+2\eta E[y^{(t+1)}f(x^{(t+1)};w_i^{(t)})|{\mathcal{F}}_t]\\
&\le&{\|w_i^{(t)}\|}_2^2+{\eta}^2{(1+2\epsilon)}^2
\end{eqnarray*}

The second inequality is because that

\begin{eqnarray*}
E[y^{(t+1)}f(x^{(t+1)};w_i^{(t)})|{\mathcal{F}}_t]&=&\frac{1}{2}E[f(x^{(t+1)};w_i^{(t)})|y^{(t+1)}=1,{\mathcal{F}}_t]\\
&-&\frac{1}{2}E[f(x^{(t+1)};w_i^{(t)})|y^{(t+1)}=-1,{\mathcal{F}}_t]\\
&\le&0
\end{eqnarray*}

Combine with ${\|w_i^{(t)}\|}_2^2={({\alpha}_i^{(t)})}^2+{\|w_{\perp,i}^{(t)}\|}_2^2$, we have that

\begin{eqnarray*}
E[{\|w_{\perp,i}^{(t+1)}\|}_2^2|\mathcal{F}_t]\le E[{\|w_{\perp,i}^{(t)}\|}_2^2|\mathcal{F}_t]+{\eta}^2{(1+2\epsilon)}^2+{({\alpha}_i^{(t)})}^2-{({\alpha}_i^{(t+1)})}^2
\end{eqnarray*}

So we have that

\begin{eqnarray*}
E({\|w_{\perp,i}^{(t)}\|}_2^2)&\le& {\|w_{\perp,i}^{(0)}\|}_2^2+{\eta}^2{(1+2\epsilon)}^2t\\
&+&\sum_{s:{\alpha}_i^{(s)}<0,s=0\ or\ {\alpha}_i^{(s)}<0,{\alpha}_i^{(s-1)}>0}{({\alpha}_i^{(s)})}^2\\
&\le&{\|w_{\perp,i}^{(0)}\|}_2^2+{({\alpha}_i^{(0)})}^2+2{\eta}^2{(1+2\epsilon)}^2t
\end{eqnarray*}

So we have that

\begin{eqnarray*}
E({\|w_{\perp,i}^{(t)}\|}_2)&\le&\sqrt{E({\|w_{\perp,i}^{(t)}\|}_2^2)}\\
&=&\sqrt{{\|w_{\perp,i}^{(0)}\|}_2^2+{({\alpha}_i^{(0)})}^2+2{\eta}^2{(1+2\epsilon)}^2t}\\
&\le&{\|w_{\perp,i}^{(0)}\|}_2+|{\alpha}_i^{(0)}|+\sqrt{2}\eta(1+2\epsilon)\sqrt{t}
\end{eqnarray*}

According to Markov's Inequaliy, we have that for any $\sigma>0$

\begin{eqnarray*}
P\{{\|w_{\perp,i}^{(t)}\|}_2\ge t^{\frac{1}{2}+\sigma}\}&\le&\frac{{\|w_{\perp,i}^{(0)}\|}_2+|{\alpha}_i^{(0)}|+\sqrt{2}\eta(1+2\epsilon)\sqrt{t}}{t^{\frac{1}{2}+\sigma}}\\
&\le&(2+\sqrt{2}\eta(1+2\epsilon))t^{-\sigma}
\end{eqnarray*}

(c) Next we will prove the convergence of $w_i^{(t)}$.

Without loss of generality, we assume that $a_i=1$.

Similar to theorem1(1)(c), we can prove that

\begin{eqnarray*}
P\{\sin{\theta}_t\le\frac{t^{\frac{1}{2}+\sigma}}{{\alpha}_i^{(0)}+2^{-2k-1}\eta t}\}\ge1-\frac{2^{2k+1}-4}{t}-(2+\sqrt{2}\eta(1+2\epsilon))t^{-\sigma}
\end{eqnarray*}

(d) Finally, we will analyze the classification accuracy of the learned CNN over $\mathcal{D}_{\epsilon}$.

If $|\{i|a_i=1\}|=|\{i|a_i=-1\}|$, let $a_{j_1}=a_{j_2}=...=a_{j_{\frac{h}{2}}}=1$, $a_{k_1}=a_{k_2}=...=a_{k_{\frac{h}{2}}}=-1$. If $\alpha_{j_m}^{(t)}>\epsilon|\alpha_{k_m}^{(t)}|+\|w_{\perp,k_m}^{(t)}\|$, $-\alpha_{k_m}^{(t)}>\epsilon|\alpha_{j_m}^{(t)}|+\|w_{\perp,j_m}^{(t)}\|$, $\forall 1\leq m\leq \frac{h}{2}$, the network will get accuracy 1 over the $\epsilon$-noisy distribution $\mathcal{D}_{\epsilon}$.

Given $m$, $P(\alpha_{j_m}^{(t)}>\epsilon|\alpha_{k_m}^{(t)}|+\|w_{\perp,k_m}^{(t)}\|)\ge1-P(\alpha_{j_m}^{(t)}<2^{-k-1}\eta t)-P(||w_{\perp,k_m}^{(t)}||>t^{\frac{2}{3}})=1-\mathcal{O}(t^{-1})-\mathcal{O}(t^{-\frac{1}{6}})$, when $t$ is sufficiently large. (note that when ${\|w_{\perp,k_m}\|}\le t^{\frac{2}{3}}$ and $t$ is sufficiently large, we have that $\epsilon|\alpha_{k_m}^{(t)}|+\|w_{\perp,k_m}^{(t)}\|\le\epsilon(1+\eta t)+t^{\frac{2}{3}}\le2^{-k-1}\eta t$)

For the same reason, $P(-\alpha_{k_m}^{(t)}>\epsilon|\alpha_{j_m}^{(t)}|+||w_{\perp,j_m}^{(t)}||)=1-\mathcal{O}(t^{-1})-\mathcal{O}(t^{-\frac{1}{6}})$, when $t$ is sufficiently large.

Thus, with any large probability, when $t$ is sufficiently large, the network will get accuracy 1 over the $\epsilon$-noisy distribution $\mathcal{D}_{\epsilon}$.

\paragraph{Theorem 3}
(\textbf{Empirical Loss from Clean Data Distribution})
Assumed that $p_+=-p_-=p$, $(x^{(t+1)},y^{(t+1)}),t=0,1,...$ are uniformly i.i.d. drawn from the training set $S=\{(x_i,y_i)\}_{i=1}^{| S|}$ where $S$ is an empirical version of the clean distribution $\mathcal{D}$. Then

(1) There exists $\mu>0$, Let $\mathcal{A}_1$: $\forall w_{\perp}\in\{w_{\perp}:\langle p,w_{\perp}\rangle=0\}$, $P_{(x,y)\sim S}\{y=1,f(x;w_{\perp})=0\}\ge\mu$, $P_{(x,y)\sim S}\{y=-1,f(x;w_{\perp})=0\}\ge\mu$, then $P(\mathcal{A}_1)=1-\mathcal{O}(\frac{1}{|S|})$.

(2) For any $\epsilon>0$, Let $\mathcal{A}_2$: $\forall w_{\perp}\in\{w_{\perp}:\langle p,w_{\perp}\rangle=0,{\|w_{\perp}\|}_2=1\}$, $|E_{(x,y)\sim S}[yf(x;w_{\perp}]|\le\epsilon$, then $P(\mathcal{A}_2)=1-\mathcal{O}(\frac{1}{|S|\epsilon^{d+1}})$.

(3.1) Let ${\theta}_t$ be the angle between $a_ip$ and $w_i^{(t)}$, if $\mathcal{A}_1$ and $\mathcal{A}_2$ hold, then $\forall K>1$, $P(\sin{\theta}_t<\frac{4K\eta}{\mu}\epsilon)\rightarrow\frac{1}{K}$ when $t\rightarrow\infty$.

(3.2) If $\mathcal{A}_1$ and $\mathcal{A}_2$ hold and $|\{i|a_i=1\}|=|\{i|a_i=-1\}|$, the classification accuracy over $\mathcal{D}$ will be 1 with probability at least $1-\mathcal{O}(\frac{\eta h}{\mu}\epsilon)$ when $t\rightarrow\infty$.

\paragraph{Proof}Without loss of generality, we assume that $u^*$ in definition1 is equaled to $1$.

%(1)We only need prove that there exists $\mu>0$, when $|S|$ is sufficiently large, then for any $w_{\perp}\in\{w_{\perp}:\langle p,w_{\perp}\rangle=0,{\|w_{\perp}\|}_2=1\}$, $P_{(x,y)\sim S_+}\{y=1,f(x;w_{\perp})=0\}\ge\mu$ and $P_{(x,y)\sim S_-}\{y=-1,f(x;w_{\perp})=0\}\ge\mu$.

(1) Let $S_+$, $S_-$ be positive examples and negative examples. Let $\phi(x,w_{\perp})=max\{\langle x_{d+1\dots 2d},w_{\perp}\rangle,\langle x_{2d+1\dots 3d}\},w_{\perp}\rangle,\dots,\langle x_{D-d+1\dots D}\},w_{\perp}\rangle\}$, where $||x||=||w_{\perp}||=1$, then $\phi(x,w_{\perp})$ is a continuous function over a bounded close set, so it is an uniformly continuous function.

So for any $\delta>0$, there exists ${\delta}_1>0$, when ${\|(x,w_{\perp})-(\hat{x},\hat{w_{\perp}})\|}_2\le{\delta}_1$, then $|\phi(x,w_{\perp})-\phi(\hat{x},\hat{w_{\perp}})|\le\delta$.

And for ${\delta}_1$, there exist a integer $M=\mathcal{O}(\frac{1}{\delta_1^{d-1}})$ and $w_{\perp,1},...,w_{\perp,M}\in\{w_{\perp}:\langle p,w_{\perp}\rangle=0,{\|w_{\perp}\|}_2=1\}$ satisfy that for any $w_{\perp}\in\{w_{\perp}:\langle p,w_{\perp}\rangle=0,{\|w_{\perp}|\}_2=1\|}$, there exists $m\in[M]$, satisfies that ${||w_{\perp}-w_{\perp,m}||}_2\le{\delta}_1$.

According to the definition of $\Dcal$, there exists $\mu>0$, for any $w_{\perp}\in\{w_{\perp}:\langle p,w_{\perp}\rangle=0,{\|w_{\perp}\|}_2=1\}$, we have that $P_{(x,y)\sim\Dcal, y=1}\{\phi(x;w_{\perp})<-\delta\}=8\mu$ and $P_{(x,y)\sim\Dcal, y=-1}\{\phi(x;w_{\perp})<-\delta\}=8\mu$.

Given $w_m$, $\mathbbm{1}_{\phi(x;w_{\perp,m})=0}$ has mean $8\mu$ and variance $8\mu(1-8\mu)$. When there are $|S_+|$ positive examples, $P_{(x,y)\sim S_+}(\phi(x;w_{\perp,m})=0)=\frac{\sum_{(x,y)\sim S_+}\mathbbm{1}_{\phi(x;w_{\perp,m})=0}}{|S_+|}$ has mean $8\mu$ and variance $\frac{8\mu(1-8\mu)}{|S_+|}$.

So $P_{(x,y)\sim S_+}(\phi(x;w_{\perp,m})=0)\geq 4\mu$ holds with probability at least $1-\frac{8\mu(1-8\mu)}{(4\mu)^2 |S_+|}=1-\mathcal{O}(\frac{1}{|S_+|})$.

And $P(|S_+|>\frac{|S|}{4})=1-\mathcal{O}(\frac{1}{|S|})$, so $P_{(x,y)\sim S}(y=1,\phi(x;w_{\perp,m})=0)\geq \mu$ holds with probability at least $1-\mathcal{O}(\frac{1}{|S|})$.

For the same reason, $P_{(x,y)\sim S}(y=-1,\phi(x;w_{\perp,m})=0)\geq \mu$ holds with probability at least $1-\mathcal{O}(\frac{1}{|S|})$.

Then $P_{(x,y)\sim S}(y=1,\phi(x;w_{\perp,m})=0)\geq \mu$ and $P_{(x,y)\sim S}(y=-1,\phi(x;w_{\perp,m})=0)\geq \mu$ hold $\forall m$ with probability at least $1-\mathcal{O}(\frac{M}{|S|})=1-\mathcal{O}(\frac{1}{\delta_1^{d-1}|S|})=1-\mathcal{O}(\frac{1}{|S|})$.
%So according to the law of large numbers, when $|S|$ is sufficiently large, we have that

%\begin{eqnarray*}
%P_{(x,y)\sim\Dcal}\{y=1,\phi(x;w_{\perp})<-\delta\}&\ge&\mu,\ \forall m\in[M]\\
%P_{(x,y)\sim\Dcal}\{y=-1,\phi(x;w_{\perp})<-\delta\}&\ge&\mu,\ \forall m\in[M]\\
%\end{eqnarray*}

So for any $w_{\perp}\in\{w_{\perp}:\langle p,w_{\perp}\rangle=0,{\|w_{\perp}\|}_2=1\}$, let $m\in[M]$ be the index satisfies that ${\|(x,w_{\perp})-(x,w_{\perp,m})\|}_2\le{\delta}_1$, then we have that

\begin{eqnarray*}
P_{(x,y)\sim S}\{y=1,\phi(x;w_{\perp})\le0\}&\ge&P_{(x,y)\sim S}\{y=1,\phi(x;w_{\perp,m})\le-\delta\}
\ge\mu\\
P_{(x,y)\sim S}\{y=-1,\phi(x;w_{\perp})\le0\}&\ge&P_{(x,y)\sim S}\{y=-1,\phi(x;w_{\perp,m})\le-\delta\}
\ge\mu
\end{eqnarray*}

(2)Let $\phi(x,w_{\perp})=f(x;w_{\perp})$, then $\phi(x,w_{\perp})$ (with $||x||=||w_{\perp}||=1$) is a Lipschitz continuity function with constant $L$.

So for any $\epsilon>0$, there exists ${\epsilon}_1=\frac{\epsilon}{L}$, when ${||(x,w_{\perp})-(\hat{x},\hat{w_{\perp}})||}_2\le{\epsilon}_1$, then $|\phi(x,w_{\perp})-\phi(\hat{x},\hat{w_{\perp}})|\le\frac{\epsilon}{2}$.

And for ${\epsilon}_1$, there exist a integer $M=\mathcal{O}(\frac{1}{\epsilon_1^{d-1}})$ and $w_{\perp,1},...,w_{\perp,M}\in\{w_{\perp}:\langle p,w_{\perp}\rangle=0,{\|w_{\perp}\|}_2=1\}$ satisfy that for any $w_{\perp}\in\{w_{\perp}:\langle p,w_{\perp}\rangle=0,{\|w_{\perp}\|}_2=1\}$, there exists $m\in[M]$, satisfies that ${\|w_{\perp}-w_{\perp,m}\|}_2\le{\epsilon}_1$.

For any $m\in[M]$, we have that $|E_{(x,y)\sim\mathcal{D}}(yf(x;w_{\perp,m}))|=0$.

Given m, $yf(x;w_{\perp,m})$ is a random variable with mean $0$ and variance $\sigma^2<\infty$. So with $|S|$ examples, $E_{(x,y)\sim S}yf(x;w_{\perp,m})$ has mean $0$ and variance $\frac{\sigma^2}{|S|}$.

So $|E_{(x,y)\sim U(S)}(yf(x;w_{\perp,m}))|\le\frac{\epsilon}{2}$ holds with probability at least $1-\frac{\sigma^2}{|S|(\frac{\epsilon}{2})^2}$.
%So according to the law of large numbers, when $|S|$ is sufficiently large, we have that

And $\forall m$, $|E_{(x,y)\sim U(S)}(yf(x;w_{\perp,m}))|\le\frac{\epsilon}{2}$ hold with probability at least $1-\frac{M\sigma^2}{|S|(\frac{\epsilon}{2})^2}=1-\mathcal{O}(\frac{L^{d-1}\sigma^2}{|S|\epsilon^{d+1}})=1-\mathcal{O}(\frac{1}{|S|\epsilon^{d+1}})$.
%\begin{eqnarray*}
%|E_{(x,y)\sim U(S)}(yf(x;w_{\perp,m}))|\le\frac{\epsilon}{2},\ \forall\ m\in[M]
%\end{eqnarray*}

So for any $w_{\perp}\in\{w_{\perp}:\langle p,w_{\perp}\rangle=0,{||w_{\perp}||}_2=1\}$, let $m\in[M]$ be the index satisfies that ${||(x,w_{\perp})-(x,w_{\perp,m})||}_2\le{\epsilon}_1$, then we have that

\begin{eqnarray*}
|E_{(x,y)\sim S}(yf(x;w_{\perp}))|&=&|E_{(x,y)\sim S}(y(f(x;w_{\perp})-f(x;w_{\perp,m})))+E_{(x,y)\sim S}(yf(x;w_{\perp,m}))|\\
&\le&E_{(x,y)\sim S}(|f(x;w_{\perp})-f(x;w_{\perp,m})|)+|E_{(x,y)\sim S}(yf(x;w_{\perp,m}))|\\
&\le&\frac{\epsilon}{2}+\frac{\epsilon}{2}\\
&=&\epsilon
\end{eqnarray*}

(3)
Assumed that $|S|$ is sufficiently large so that $\mathcal{A}_1$ and $\mathcal{A}_2$ hold.

Consider the decomposition of $w_i^{(t)}$

\begin{eqnarray*}
w_i^{(t)}={\alpha}_i^{(t)}p+w_{\perp,i}^{(t)}
\end{eqnarray*}

where ${\alpha}_i^{(t)}\in \Rbb$ and $w_{\perp,i}^{(t)}$ is perpendicular to $p$.

According to Theorem3(1) and similar to Theorem1(1), we can proof that when $a_i=1$, we have that ${\alpha}_i^{(t)}$ is monotonically increasing and

\begin{eqnarray*}
P_\{{\alpha}_i^{(t)}\le{\alpha}_i^{(0)}+\frac{\mu}{2}t\}\le\frac{4(1-\mu)}{\mu t}
\end{eqnarray*}

And when $a_i=-1$, we have that ${\alpha}_i^{(t)}$ is monotonically decreasing and

\begin{eqnarray*}
P_\{{\alpha}_i^{(t)}\ge{\alpha}_i^{(0)}-\frac{\mu}{2}t\}\le\frac{4(1-\mu)}{\mu t}
\end{eqnarray*}

When $a_i=1$, if ${\alpha}_i^{(t)}>0$, the dynamics of $w_{\perp,i}^{(t)}$ is

\begin{eqnarray*}
w_{\perp,i}^{(t+1)}&=&w_{\perp,i}^{(t)}+\eta x_{u_1d+1...(u_1+1)d}\Ibb_{\{y^{(t+1)}=1,f(x^{(t+1)};w_{\perp,i}^{(t)})>{\alpha}_i^{(t)}\}}\\
&-&\eta x_{u_2d+1...(u_2+1)d}\Ibb_{\{y^{(t+1)}=-1,f(x^{(t+1)};w_{\perp,i}^{(t)})>0\}}\\
\end{eqnarray*}

where $u_1,u_2\in[k]$ and $u_1,u_2\ne1$.

So we have that

\begin{eqnarray*}
E[{\|w_{\perp,i}^{(t+1)}\|}_2^2|\mathcal{F}_t]&\le&{\|w_{\perp,i}^{(t)}\|}_2^2+{\eta}^2+2\eta E[f(x^{(t+1)};w_{\perp,i}^{(t)})\Ibb_{\{y^{(t+1)}=1,f(x^{(t+1)};w_{\perp,i}^{(t)})>{\alpha}_i^{(t)}\}}|\mathcal{F}_t]\\
&-&2\eta E[f(x^{(t+1)};w_{\perp,i}^{(t)})\Ibb_{\{y^{(t+1)}=-1,f(x^{(t+1)};w_{\perp,i}^{(t)})>0\}}|\mathcal{F}_t]\\
&\le&{\|w_{\perp,i}^{(t)}\|}_2^2+{\eta}^2+2\eta E[f(x^{(t+1)};w_{\perp,i}^{(t)})\Ibb_{\{y^{(t+1)}=1,f(x^{(t+1)};w_{\perp,i}^{(t)})>0\}}|\mathcal{F}_t]\\
&-&2\eta E[f(x^{(t+1)};w_{\perp,i}^{(t)})\Ibb_{\{y^{(t+1)}=-1,f(x^{(t+1)};w_{\perp,i}^{(t)})>0\}}|\mathcal{F}_t]\\
&=&{\|w_{\perp,i}^{(t)}\|}_2^2+{\eta}^2+2\eta E[y^{(t+1)}f(x^{(t+1)};\frac{w_{\perp,i}^{(t)}}{{\|w_{\perp,i}^{(t)}\|}_2})|\mathcal{F}_t]{\|w_{\perp,i}^{(t)}\|}_2\\
&\le&{\|w_{\perp,i}^{(t)}\|}_2^2+{\eta}^2+2\eta\epsilon{\|w_{\perp,i}^{(t)}\|}_2\\
&\le&{\|w_{\perp,i}^{(t)}\|}_2^2+{\eta}^2+1+2\eta\epsilon{\|w_{\perp,i}^{(t)}\|}_2
\end{eqnarray*}

The third inequality is according to Theorem3(2).

If ${\alpha}_i^{(t)}<0$, according to the dynamics of $w_i^{(t)}$, we have that

\begin{eqnarray*}
E[{\|w_i^{(t+1)}\|}_2^2|\mathcal{F}_t]&\le&{\|w_i^{(t)}\|}_2^2+{\eta}^2+2\eta E[y^{(t+1)}f(x^{(t+1)};w_{\perp,i}^{(t)})|\mathcal{F}_t]\\
&=&{\|w_i^{(t)}\|}_2^2+{\eta}^2+2\eta E[max\{0,f(x^{(t+1)};w_{\perp,i}^{(t)})\}\Ibb_{\{y^{(t+1)}=1\}}|\mathcal{F}_t]\\
&-&2\eta E[max\{-{\alpha}_i^{(t)},f(x^{(t+1)};w_{\perp,i}^{(t)})\}\Ibb_{\{y^{(t+1)}=-1\}}|\mathcal{F}_t]\\
&\le&{\|w_i^{(t)}\|}_2^2+{\eta}^2+2\eta E[f(x^{(t+1)};w_{\perp,i}^{(t)})\Ibb_{\{y^{(t+1)}=1\}}|\mathcal{F}_t]\\
&-&2\eta E[f(x^{(t+1)};w_{\perp,i}^{(t)})\Ibb_{\{y^{(t+1)}=-1\}}|\mathcal{F}_t]\\
&=&{\|w_i^{(t)}\|}_2^2+{\eta}^2+2\eta E[y^{(t+1)}f(x^{(t+1)};\frac{w_{\perp,i}^{(t)}}{{\|w_{\perp,i}^{(t)}\|}_2})|\mathcal{F}_t]{\|w_{\perp,i}^{(t)}\|}_2\\
&\le&{\|w_i^{(t)}\|}_2^2+{\eta}^2+2\eta\epsilon{\|w_{\perp,i}^{(t)}\|}_2
\end{eqnarray*}

Combine with ${\|w_i^{(t)}\|}_2^2={({\alpha}_i^{(t)})}^2+{\|w_{\perp,i}^{(t)}\|}_2^2$, we have that

\begin{eqnarray*}
E[{\|w_{\perp,i}^{(t+1)}\|}_2^2|\mathcal{F}_t]&\le& {\|w_{\perp,i}^{(t)}\|}_2^2+{\eta}^2+2\eta\epsilon{\|w_{\perp,i}^{(t)}\|}_2+{({\alpha}_i^{(t)})}^2-{({\alpha}_i^{(t+1)})}^2\\
&\le&{\|w_{\perp,i}^{(t)}\|}_2^2+{\eta}^2+1+2\eta\epsilon{\|w_{\perp,i}^{(t)}\|}_2
\end{eqnarray*}

So no matter what ${\alpha}_i^{(t)}$ is, we always have that

\begin{eqnarray*}
E({\|w_{\perp,i}^{(t+1)}\|}_2^2)&\le&E({\|w_{\perp,i}^{(t)}\|}_2^2)+{\eta}^2+1+2\eta\epsilon E({\|w_{\perp,i}^{(t)}\|}_2)\\
&\le&E({\|w_{\perp,i}^{(t)}\|}_2^2)+{\eta}^2+1+2\eta\epsilon\sqrt{E({\|w_{\perp,i}^{(t)}\|}_2^2)}\\
\end{eqnarray*}

Let $e_0={(\frac{{\eta}^2+1}{2\eta\epsilon})}^2$ and $e_{t+1}=e_t+{\eta}^2+1+2\eta\epsilon\sqrt{e_t}$, then $\{e_t\}$ is monotonically increasing so that $e_t\ge{(\frac{{\eta}^2+1}{2\eta\epsilon})}^2$ always holds.

Without loss of generality we assumed that $e_0\ge1$(only need $\epsilon\leq 1$), so that $E({\|w_{\perp,i}^{(t)}\|}_2^2)\le e_t$ always holds.

According to the dynamics of $e_t$, we have that

\begin{eqnarray*}
e_{t+1}&=&e_t+{\eta}^2+1+2\eta\epsilon\sqrt{e_t}\\
&\le&e_t+4\eta\epsilon\sqrt{e_t}\\
&\le&{(\sqrt{e_t}+2\eta\epsilon)}^2
\end{eqnarray*}

So we have that $\sqrt{e_{t+1}}\le\sqrt{e_t}+2\eta\epsilon$ and so that $\sqrt{e_t}\le\sqrt{e_0}+2\eta\epsilon t$.

So we have that

\begin{eqnarray*}
E({\|w_{\perp,i}^{(t)}\|}_2)&\le&\sqrt{e_t}\\
&\le&\sqrt{e_0}+2\eta\epsilon t\\
&=&\frac{{\eta}^2+1}{2\eta\epsilon}+2\eta\epsilon t
\end{eqnarray*}

For any $K>1$, according to Markov's inequality, we have that

\begin{eqnarray*}
P\{{\|w_{\perp,i}^{(t)}\|}_2\ge 2\eta\epsilon tK\}\le\frac{\frac{{\eta}^2+1}{2\eta\epsilon}+2\eta\epsilon t}{2\eta\epsilon tK}
\end{eqnarray*}

So when $a_i=1$, $P(\sin{\theta_t}\geq\frac{4K\eta\epsilon}{\mu})>P(\tan{\theta_t}>\frac{4K\eta\epsilon}{\mu})\geq 1-P\{{\|w_{\perp,i}^{(t)}\|}_2\ge 2\eta\epsilon tK\}-P(\alpha_i^{(t)}<\frac{\mu t}{2})\rightarrow \frac{1}{K}$, while $t\rightarrow\infty$.

If $|\{i|a_i=1\}|=|\{i|a_i=-1\}|$, let $a_{j_1}=a_{j_2}=...=a_{j_{\frac{h}{2}}}=1$, $a_{k_1}=a_{k_2}=...=a_{k_{\frac{h}{2}}}=1$. If $\alpha_{j_m}^{(t)}>\|w_{k_m,\vert}^{(t)}\|$, $-\alpha_{k_m}^{(t)}>\|w_{j_m,\vert}^{(t)}\|$, $\forall 1\leq m\leq \frac{h}{2}$, the network will get accuracy 1 over the clean distribution $\mathcal{D}$.

Given $m$, $P(\alpha_{j_m}^{(t)}>\|w_{k_m,\vert}^{(t)}\|)\ge1-P(\alpha_{j_m}^{(t)}<\frac{\mu t}{2})-P(||w_{k_m,\vert}^{(t)}||>\frac{\mu t}{2})=1-\mathcal{O}(\frac{\eta}{\mu}\epsilon)$ when $t\rightarrow\infty$.

For the same reason, $P(-\alpha_{k_m}^{(t)}>||w_{j_m,\vert}^{(t)}||)=1-\mathcal{O}(\frac{\eta}{\mu}\epsilon)$ when $t\rightarrow\infty$.

Thus, the network will get accuracy 1 over the clean distribution $\mathcal{D}$ with probability at least $1-\mathcal{O}(\frac{\eta h}{\mu}\epsilon)$.

\subsection*{Proofs for Section 4}

\paragraph{Theorem 4}
(\textbf{Learning with Multiple Key Patterns}) 
Assumed that $(x^{(t+1)},y^{(t+1)}),t=0,1,...$ are i.i.d. drawn from the multi pattern clean distribution $\mathcal{D}_{mul}$ and $p_{+,1}$, $p_{+,2}$, $p_{-,1}$, $p_{-,2}$ are perpendicular to each other. Then

(1) When $a_i=1$, if $\langle p_{+,1},w_i^{(0)}\rangle>\langle p_{+,2},w_i^{(0)}\rangle$, let ${\theta}_t$ be the angle between $p_{+,1}$ and $w_i^{(t)}$, if $\langle p_{+,2},w_i^{(0)}\rangle>\langle p_{+,1},w_i^{(0)}\rangle$, let ${\theta}_t$ be the angle between $p_{+,2}$ and $w_i^{(t)}$. If $max\{\langle p_{+,1},w_i^{(0)}\rangle,\langle p_{+,2},w_i^{(0)}\rangle\}>0$ hold, then

\begin{equation}
P\{\sin{\theta}_t\le \mathcal{O}(t^{-(\frac{1}{2}-\sigma)})\}\ge1-\mathcal{O}(t^{-\sigma})
\end{equation}

(2) When $a_i=-1$, if $\langle p_{-,1},w_i^{(0)}\rangle>\langle p_{-,2},w_i^{(0)}\rangle$, let ${\theta}_t$ be the angle between $p_{-,1}$ and $w_i^{(t)}$, if $\langle p_{-,2},w_i^{(0)}\rangle>\langle p_{-,1},w_i^{(0)}\rangle$, let ${\theta}_t$ be the angle between $p_{-,2}$ and $w_i^{(t)}$. If $max\{\langle p_{-,1},w_i^{(0)}\rangle,\langle p_{-,2},w_i^{(0)}\rangle\}>0$ hold, then

\begin{equation}
P\{\sin{\theta}_t\le \mathcal{O}(t^{-(\frac{1}{2}-\sigma)})\}\ge1-\mathcal{O}(t^{-\sigma})
\end{equation}

\paragraph{Proof}
Consider the decomposition of $w_i^{(t)}$

\begin{eqnarray*}
w_i^{(t)}=w_{\parallel,+,i}^{(t)}+w_{\parallel,-,i}^{(t)}+w_{\perp,i}^{(t)}
\end{eqnarray*}

where $w_{\parallel,+,i}^{(t)}\in span\{p_{+,1},p_{+,2}\}$, $w_{\parallel,-,i}^{(t)}\in span\{p_{-,1},p_{-,2}\}$, $w_{\perp,i}^{(t)}$ is perpendicular to $p_{+,1}$, $p_{+,2}$, $p_{-,1}$, $p_{-,2}$.

Without loss of generality, we assume that $a_i=1$

According to the definition of the multi pattern clean distribution $\Dcal_{mul}$, if $\langle p_{+,1},w_i^{(0)}\rangle>max\{0,\langle p_{+,2},w_i^{(0)}\rangle\}$, then $w_i^{(t)}$ will not activate $p_{+,2}$ and with probability at least $2^{-k+1}$, in step $t$, $w_i^{(t)}$ will activate $p_{+,1}$. If $\langle p_{+,2},w_i^{(0)}\rangle>max\{0,\langle p_{+,1},w_i^{(0)}\rangle\}$, then $w_i^{(t)}$ will not activate $p_{+,1}$ and with probability at least $2^{-k+1}$, in step $t$, $w_i^{(t)}$ will activate $p_{+,2}$.

According to the dynamics of $w_{\parallel,-,i}^{(t)}$, we can prove that the norm of $w_{\parallel,-,i}^{(t)}$ will not be larger than $max\{{\|w_{\parallel,-,i}^{(0)}\|}_2,\sqrt{2}+\sqrt{2}\eta\}$.

Similar to Theorem 1(2), for any $\sigma\in(0,\frac{1}{2})$, we have that

\begin{eqnarray*}
P\{{\|w_{\perp,i}^{(t)}\|}_2\ge\mathcal{O}(t^{\frac{1}{2}+\sigma})\}\le\mathcal{O}(t^{-\sigma})
\end{eqnarray*}

So we have that for any $\sigma\in(0,\frac{1}{2})$, if $max\{\langle p_{-,1},w_i^{(0)}\rangle,\langle p_{-,2},w_i^{(0)}\rangle\}>0$ hold, then

\begin{eqnarray*}
P\{\sin{\theta}_t\le \mathcal{O}(t^{-(\frac{1}{2}-\sigma)})\}\ge1-\mathcal{O}(t^{-\sigma})
\end{eqnarray*}

\paragraph{Theorem 5}
Assumed that every mini-batch is i.i.d. randomly drawn from the clean distribution $\mathcal{D}$ with as many positive examples as negative examples, and $p_+=-p_-=p$, with initialization $a_i^{(0)}=\pm 1$, and $||w_i^{(0)}||<\eta \ll 1$, let $w_i^{(t)}=\alpha_i^{(t)}p+w_{i,\perp}^{(t)}$, then $sgn(a_i^{(t)})=sgn(a_i^{(0)})$, $a_i^{(t)}\rightarrow\infty$ and for $t>\frac{1}{\eta}$, $\alpha_i^{(t)}=sgn(a_i^{(0)})$,$\alpha_i^{(t)}>||w_{i,\perp}^{(t)}||$, hold with probability $p=1-\mathcal{O}(\eta)$.

%\section{proof of theorem 5}
\paragraph{Proof}
Given $0<\delta<1$, assume when $1\leq t\leq K$,
\begin{equation}
|a_i^{(t)}-a_i^{(0)}|<\delta |a_i^{(0)}|
\label{aueq}
\end{equation}

then
\begin{equation}
\|w_i^{(t)}\|<\|w_i^{(0)}\|+(1+\delta)|a_i^{(0)}|\eta t.
\label{wueq}
\end{equation}

Since $|a_i^{(t)}-a_i^{(t-1)}|<\eta \|w_i^{(t-1)}\|$,
when
\begin{equation}
\eta K(\|w_i^{(0)}\|+(1+\delta)|a_i^{(0)}|\eta K)<\delta |a_i^{(0)}|
\label{kueq}
\end{equation}
(\ref{aueq}), (\ref{wueq}) hold.

With initialization $a_i^{(0)}\sim U(\{-1,1\})$, and $||w_i^{(0)}||<\eta \ll 1$, when $K<\frac{1}{\eta}\sqrt{\frac{\delta}{1+\delta}}-1$,
\begin{equation}
\begin{aligned}
&\eta K(||w_i^{(0)}||+(1+\delta)|a_i^{(0)}|\eta K)\\
<&\eta(K+1)(\eta(1+\delta)+(1+\delta)\eta K)\\
<&\eta^2(K+1)^2(1+\delta)\\
<&\delta=\delta |a_i^{(0)}|
\end{aligned}
\end{equation}
Thus, (\ref{kueq}) holds.

After $K$ steps, $|a_i^{(t)}-a_i^{(0)}|<\delta |a_i^{(0)}|$, so $sgn(a_i^{(t)})=sgn(a_i^{(0)})$.

Without loss of generality, we assume $a_i^{(0)}=1$.

$w_i^{(K)}=w_{i,\parallel}^{(K)}+w_{i,\perp}^{(K)}=\alpha_i^{(K)}p+w_{i,\perp}^{(K)}$, as in proof of Theorem 1, $\alpha_i^{(K)}\geq \alpha_i^{(0)}+(1-\delta)\sum_{t=1}^{K} e_t$  and $E||w_{i,\perp}^{(K)}||^2\leq E||w_{i,\perp}^{(0)}||^2+E||w_{i,\parallel}^{(0)}||^2+K\eta^2(1+\delta)$, since $|\alpha_i^{(0)}|<||w_i^{(0)}||<\eta$, then with probability $P_k=1-\mathcal{O}(\frac{K\eta^2(1+\delta)}{K^2\eta^2p_0^2(1-\delta)^2})$,  $\alpha_i^{(K)}>||w_{i,\perp}^{(K)}||$.

Let $K=\frac{1}{\eta}\sqrt{\frac{\delta}{1+\delta}}-2$, so with probability $p=1-\mathcal{O}(\eta)$, $\alpha_i^{(K)}>||w_{i,\perp}^{(K)}||$.

When $\alpha_i^{(K)}>||w_{i,\perp}^{(K)}||$ and there are as many positive examples as negative examples in every mini-batch, $\forall t>K$, $a_i^{(t)}$ and $\alpha_i^{(t)}$ monotonic nondecreasing and $\alpha_i^{(t)}>||w_{i,\perp}^{(t)}||$ holds (since every key pattern in positive samples will all be actived). So when $t>\frac{1}{\eta}>K$, the conclusion holds.
\end{document}